\documentclass[letterpaper, preprint, paper,11pt]{AAS}	

\usepackage{bm}
\usepackage{amsmath}
\usepackage{subcaption}
\usepackage[colorlinks=true, pdfstartview=FitV, linkcolor=black, citecolor= black]{hyperref}
\usepackage{overcite}
\usepackage{footnpag}			      	
\usepackage{adjustbox}
\usepackage[most]{tcolorbox}
\usepackage{colortbl}
\usepackage[normalem]{ulem}
\usepackage{makecell}

\usepackage{booktabs} 
\newcommand{\ra}[1]{\renewcommand{\arraystretch}{#1}} 
\definecolor{aureolin}{rgb}{0.99, 0.93, 0.0}
\definecolor{aqua}{rgb}{0.0, 1.0, 1.0}
\definecolor{blue-violet}{rgb}{0.54, 0.17, 0.89}

\usepackage{graphicx}

\newcommand{\vvec}[1]{\bm{\mathrm{#1}}}

\usepackage{enumitem,amssymb}
\newlist{todolist}{itemize}{2}
\setlist[todolist]{label=$\square$}
\usepackage{pifont}
\def\thefootnotee{$\star$}\footnotetext{$\star$These authors contributed equally to this work.}

\PaperNumber{23-294}

\begin{document}

\title{Deep Monocular Hazard Detection for\protect\\ Safe Small Body Landing}

\author{
Travis Driver\thefootnotee{}\thanks{PhD Student, Institute for Robotics and Intelligent Machines, School of Aerospace Engineering, Georgia Institute of Technology, Atlanta, GA 30332, USA.}, 
\ Kento Tomita\thefootnotee{}\thanks{PhD Student, School of Aerospace Engineering, Georgia Institute of Technology, Atlanta, GA 30332, USA.},
\ Koki Ho\thanks{Associate Professor, School of Aerospace Engineering, Georgia Institute of Technology, Atlanta, GA 30332, USA.},
\ and Panagiotis Tsiotras\thanks{David \& Andrew Lewis Chair, Professor, Institute for Robotics and Intelligent Machines, School of Aerospace Engineering, Georgia Institute of Technology, Atlanta, GA 30332, USA.}
}

\maketitle{}

\begin{abstract}
Hazard detection and avoidance is a key technology for future robotic small body sample return and lander missions. 
Current state-of-the-practice methods rely on high-fidelity, \textit{a priori} terrain maps, which require extensive human-in-the-loop verification and expensive reconnaissance campaigns to resolve mapping uncertainties. 
We propose a novel safety mapping paradigm that leverages deep semantic segmentation techniques to predict landing safety directly from a single monocular image, thus reducing reliance on high-fidelity, \textit{a priori} data products. 
We demonstrate precise and accurate safety mapping performance on real \textit{in-situ} imagery of prospective sample sites from the OSIRIS-REx mission.
\end{abstract}


\section{Introduction}
Hazard detection and avoidance (HD\&A) is a key technology for future robotic small body sample return and lander missions. 
Current approaches rely on high-fidelity digital elevation maps (DEMs) derived from digital terrain models (DTMs), local topography and albedo maps, generated on the ground~\cite{berry2022scitech}. 
However, DTM construction involves extensive human-in-the-loop verification, carefully designed image acquisition plans, and expensive reconnaissance campaigns to resolve mapping uncertainties~\cite{barnouin2020,palmer2022practical}. 
We, instead, propose a novel safety mapping paradigm that leverages Bayesian deep learning techniques to accurately predict landing safety maps \textit{directly from monocular images} in order to reduce reliance on expensive high-fidelity, \textit{a priori} data products (i.e., DTMs). 

Safety mapping methodologies that leverage deep learning have demonstrated potential to improve the accuracy of onboard hazard detection. 
Previous works~\cite{moghe2020jsr,tomita2022jsr} have leveraged deep semantic segmentation to classify safe and unsafe landing locations from digital elevation maps (DEMs) derived from simulated LiDAR scans. 
However, generating reliable DEMs from LiDAR scans is non-trivial and requires accurate state estimates and range measurements. 
Moreover, LiDARs typically feature a relatively small effective operating range~\cite{lorenz2017} and increased size, weight, and power (SWaP) requirements relative to passive sensors such as monocular cameras.
Thus, we propose to derive landing safety maps directly from monocular images without assuming any \textit{a priori} data or relying on the fidelity of the current state estimate. 
Deep semantic segmentation has been previously employed for surface characterization of small bodies from simulated monocular images, primarily focusing on boulder detection~\cite{pugliatti2022jsr,caroselli2022gnc}. 
Conversely, we apply our models to real images and directly predict safety maps that conform to realistic landing parameters and constraints. 

\newpage

The contributions of this paper are as follows: 
first, we develop a novel safety mapping paradigm that leverages Bayesian deep learning techniques to predict landing safety \textit{directly from monocular images}; 
second, we construct a dataset of \textit{real} monocular images and corresponding landing safety maps that conform to realistic landing parameters for training and testing our models; 
third, we demonstrate precise and accurate safety mapping performance on real imagery of prospective sample sites from the recent OSIRIS-REx mission to Asteroid 101955 Bennu. 
Our code, data, and trained models will be made available to the public at \url{https://github.com/travisdriver/deep_monocular_hd}.


\section{Related Work}

Current hazard detection methodologies for small body missions rely on high-fidelity digital elevation maps (DEMs) derived from digital terrain models (DTMs), local topography and albedo maps~\cite{berry2022scitech}. 
However, DTM construction typically involves extensive human-in-the-loop verification and carefully designed image acquisition plans to achieve optimal results~\cite{barnouin2020,palmer2022practical}.
Consequently, autonomous hazard detection and avoidance (HD\&A) has been identified as a high-priority technology~\cite{nasa2020tech} to promote and enable new mission concepts to near-earth asteroids, comets, the Moon, Mars, and beyond. 

The Autonomous Landing Hazard Avoidance Technology (ALHAT)~\cite{epp2008alhat,carson2014alhat} program was launched in 2005, followed by the Safe \& Precise Landing---Integrated Capabilities Evolution (SPLICE) program~\cite{sostaric2021splice} in 2018, in order to develop autonomous landing technologies. 
These programs have focused on developing HD\&A algorithms that operate on DEMs generated from range measurements acquired by active sensors such as flash LiDARs. 
However, these methods are constrained by the relatively small effective operating range and the increased size, weight, and power (SWaP) requirements of LiDARs relative to passive sensors such as monocular cameras. 
Indeed, the OSIRIS-REx Guidance, Navigation, and Control (GNC) flash LiDAR had a maximum operational range of approximately 1 km~\cite{church2020lidar,leonard2022lidar}, while preliminary testing of the Hazard Detection LiDAR of the SPLICE program demonstrated a 5 cm ground sample distance at 500 meters and near-nadir pointing~\cite{sostaric2021splice}. 
Conversely, the OSIRIS-REx Camera Suite (OCAMS)~\cite{rizk2018ocams} was able to acquire 5 cm GSD images at almost 4 km, providing higher-resolution measurements earlier in the mission than the active sensors onboard and allowing for detailed surface characterization during the early phases of the mission~\cite{lorenz2017}. 
Moreover, constructing a DEM from LiDAR scans is non-trivial and requires accurate range measurements and precise knowledge of the spacecraft's relative pose to the landing plane. 
Instead, we focus on estimating landing safety directly from a single monocular image. 

Safety mapping methodologies that leverage deep learning have demonstrated potential to improve hazard detection accuracy and have also been shown to offer competitive runtimes on flight relevant hardware~\cite{claudet2022benchmark}. 
Previous works have leveraged deep semantic segmentation to classify safe and unsafe landing locations from high-resolution DEMs. 
Moghe and Zanetti~\cite{moghe2020jsr} leverage a deep neural network architecture for predicting safety maps from a DEM and design a novel loss function specifically designed to decrease the false safe rate and encourage more precise safe predictions. 
Tomita et al.~\cite{tomita2022jsr} employ a Bayesian SegNet architecture~\cite{kendall2017bsegnet} for segmentation of input DEMs into safe and unsafe landing locations. 
The Bayesian architecture implemented by Tomita et al.~\cite{tomita2022jsr} enables uncertainty quantification of the predicted safety map through the predictive entropy of the model, allowing for more precise predictions through global thresholding with respect to this uncertainty measure.
We build upon this work and demonstrate its efficacy on monocular imagery. 

Methods based on deep learning have also been developed for surface segmentation from monocular imagery. 
Pugliatti and Maestrini~\cite{pugliatti2022jsr} employ a custom U-Net architecture for classification of surface landmarks, i.e., boulders and crater rims.  
Caroselli et al.~\cite{caroselli2022gnc} apply deep semantic segmentation to boulder detection on synthetic images of a fabricated small body model and post-process the network prediction to derive a landing safety map based on boulder density. 
Conversely, we apply our models to real images and directly predict safety maps that conform to realistic landing parameters and constraints. 


\section{Proposed Approach}
Our novel safety mapping paradigm leverages Bayesian deep learning techniques to develop an uncertainty-aware semantic segmentation model to predict safety maps from monocular imagery. 
We train and test our model on real \textit{in-situ} imagery from the OSIRIS-REx mission to asteroid 101955 Bennu with corresponding ground truth safety maps generated using realistic landing parameters and constraints. 


\subsection{Bayesian Deep Learning}

Given training input data $\mathcal{X} = \{\vvec{x}_1, \ldots, \vvec{x}_N\}$ with corresponding labels $\mathcal{Y} = \{y_1, \ldots, y_N\}$, Bayesian deep learning employs \textit{Bayesian inference} to maximize the posterior distribution of the network parameters $\vvec{\theta}$ given the training data $\mathcal{X}$, $\mathcal{Y}$:
\begin{equation}
    p(\vvec{\theta} \,|\, \mathcal{X}, \mathcal{Y}) = \frac{p(\mathcal{Y} \,|\, \mathcal{X}, \vvec{\theta}) p(\vvec{\theta})}{p(\mathcal{Y} \,|\, \mathcal{X})}.
\end{equation}
%
The distribution above can then be used to predict the likelihood of an output $y^*$ for a new input $\vvec{x}^*$ via 
\begin{equation} \label{eq:exact_pred}
    p(y^* \,|\, \vvec{x}^*, \mathcal{X}, \mathcal{Y}) = \mathbb{E}_{p(\vvec{\theta} \,|\, \mathcal{X}, \mathcal{Y})}[p(y^* \,|\, \vvec{x}^*, \vvec{\theta})], 
\end{equation}
where we may assume a softmax likelihood for $p(y^* \,|\, \vvec{x}^*, \vvec{\theta})$. 
However, computing $p(\vvec{\theta} \,|\, \mathcal{X}, \mathcal{Y})$ is intractable and must be approximated using \textit{variational inference}. 

Gal and Ghahramani~\cite{gal2015bayesian,gal2016dropout} showed that training a deep convolutional neural network (CNN) with dropout layers is equivalent to approximate variational inference with a variational distribution $q(\vvec{\theta})$ which imposes a Bernoulli distribution over the model weights. 
Specifically, consider a convolutional layer $i$ with $c_{i-1}$ input channels, $c_i$ output channels, and kernel size $k$. 
Then dropout can be viewed as imposing a distribution over the layer weights $W_i$ according to
\begin{equation}
    W_i = M_i \, \text{diag}([\epsilon_j]_{j=1}^{c_i}), \quad \epsilon_j \sim \text{Bern}(p_j),
\end{equation}
where $\epsilon_j$ are Bernoulli distributed random variables with parameter $p_j$ (which we take to be 0.5), and $M_i \in \mathbb{R}^{c_{i-1} \times k \times k \times c_{i}}$ are the variational weight parameters optimized during training. 
Therefore, Equation \eqref{eq:exact_pred} may be approximated by 
\begin{equation}
    p(y^* \,|\, \vvec{x}^*, \mathcal{X}, \mathcal{Y}) \approx \mathbb{E}_{q(\vvec{\theta})}[p(y^* \,|\, \vvec{x}^*, \vvec{\theta})].
\end{equation}

Finally, employing dropout at test time permits the use of the predictive entropy approximated through $T$ \textit{stochastic forward passes} through the network as a measure of uncertainty~\cite{mukhoti2018evaluating}:
%
\begin{equation} \label{eq:entropy}
    \mathbb{H}[y \,|\, \vvec{x}, \mathcal{X}, \mathcal{Y}] \approx -\sum_{k=1}^d\left(\frac{1}{T}\sum_{t=1}^T p(y=k \,|\, \vvec{x}, \vvec{\theta}_t)\right) \log \left(\frac{1}{T}\sum_{t=1}^T p(y=k \,|\, \vvec{x}, \vvec{\theta}_t)\right),
\end{equation}
%
where $\vvec{\theta}_t$ corresponds to a realization of the network parameters, distributed according to $q(\vvec{\theta})$, sampled during a forward pass through the network, and the average over the forward passes is taken to be the final prediction probabilities. 
This process is referred to as \textit{Monte Carlo (MC) dropout}~\cite{gal2015bayesian,gal2016dropout}. 
For the task of semantic segmentation, assume $\vvec{x}$ is a tuple $(\vvec{X}, \vvec{u})$ containing an image tensor $\vvec{X} \in \mathbb{R}^{h\times w\times c}$ and an image coordinate $\vvec{u} \in \mathbb{R}^2$, and $k \in \{1, \ldots, d\}$ is a \textit{pixel-wise} class label for the pixel located at $\vvec{u}$.
We will demonstrate that leveraging this uncertainty measure for our network predictions leads to increased precision and accuracy of safe landing locations.


\subsection{Uncertainty-Aware Semantic Segmentation}

\begin{figure}
    \centering
    \includegraphics[width=\linewidth]{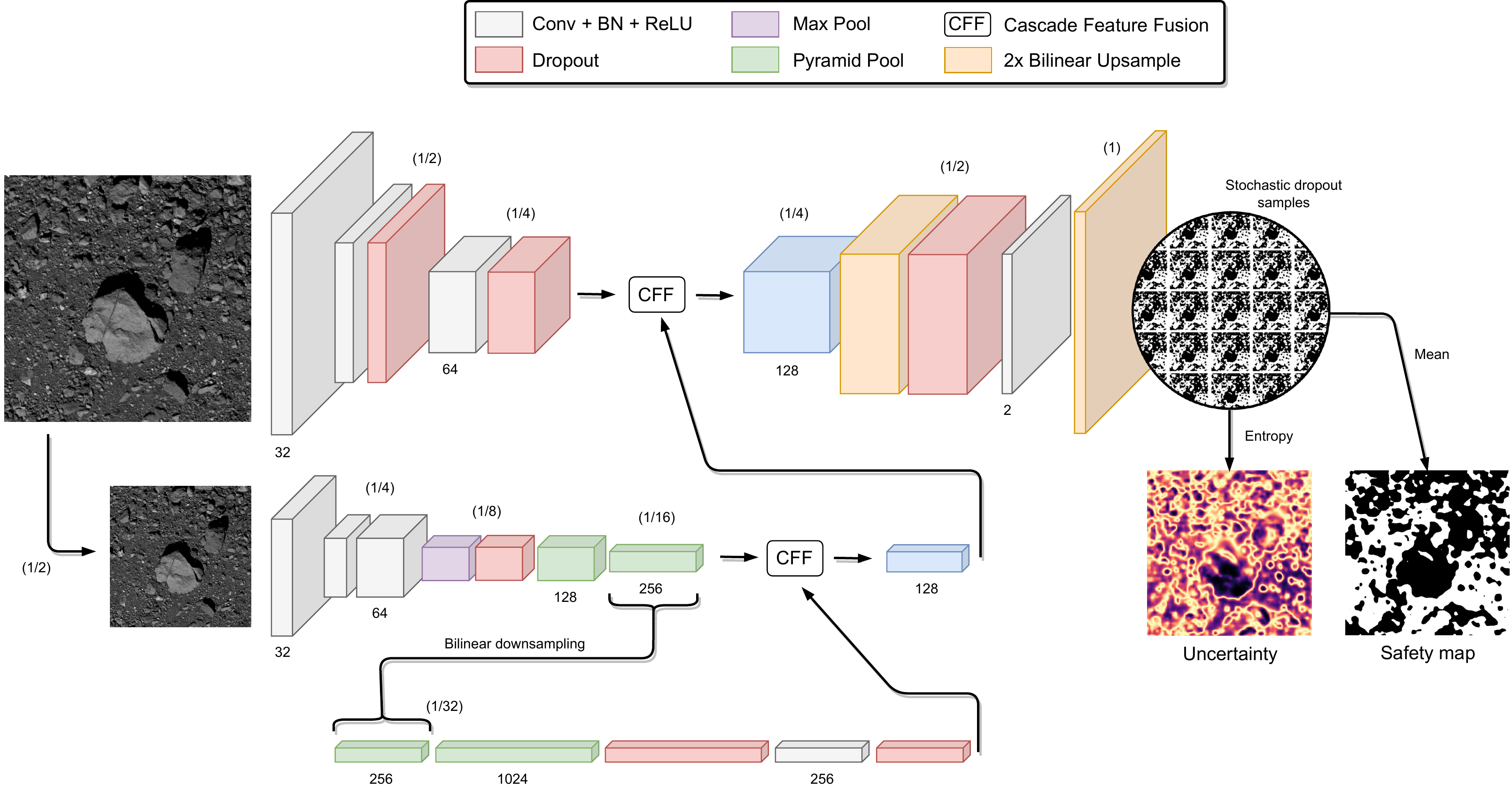}
    \caption{\textbf{Bayesian ICNet architecture.} Multiscale fusion is conducted within the cascade feature fusion (CFF)~\cite{zhao2018icnet} modules. The ratio in parentheses denotes the relative magnitude of the spatial dimensions with respect to the original image.}
    \label{fig:network-arch}
\end{figure}

We leverage an uncertainty-aware semantic segmentation architecture based on the image cascade network (ICNet)~\cite{zhao2018icnet}, shown in Figure \ref{fig:network-arch}. 
ICNet is a highly efficient segmentation architecture that blends coarse prediction maps obtained from down-sampled inputs with high-resolution feature maps obtained from high throughput networks that operate on the full-resolution image, allowing for fast inference on high-resolution images while maintaining accuracy. 
Multiscale feature map fusion is conducted by the \textit{cascade feature fusion (CFF)} modules, whereby a reduced-resolution segmentation map is computed from the two multiscale feature map inputs.  
The multiscale predictions are used to train the network via a weighted softmax cross-entropy loss~\cite{zhao2018icnet}. 

We implement a Bayesian version of ICNet, which we denote as BICNet, where dropout layers are added to allow for stochastic sampling with respect to the model parameters using techniques from Bayesian deep learning, i.e., MC dropout, as described in the previous subsection. 
Ideally, a Bayesian NN would feature a dropout layer after every hidden layer of the network~\cite{gal2015bayesian, gal2016dropout}.
However, as observed in previous works~\cite{mukhoti2018evaluating,kendall2017bsegnet}, adding dropout layers after every convolutional layer in more complex networks is too strong of a regularizer, resulting in underfitting. 
Therefore, we follow the work of Mukhoti et al.~\cite{mukhoti2018evaluating} and Kendall and Cippola~\cite{kendall2017bsegnet} and only insert dropout layers after the central encoder and decoder layers. 
At test time, we perform $T=8$ stochastic forward passes and use the predictive entropy, defined in Equation \eqref{eq:entropy}, as a measure of uncertainty, and use the average of this measure over all training instances as a threshold to mask out high uncertainty regions in the image. 


\subsection{Data Generation}

\begin{figure}[tb!]
    \centering
    \input{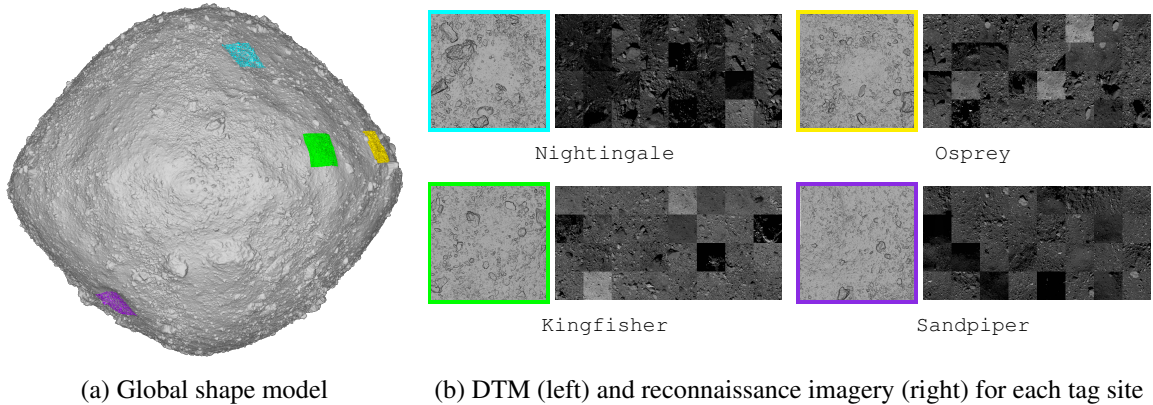}
    \caption{\textbf{OSIRIS-REx TAG site datasets.} TAG site locations are indicated by the corresponding color in the global shape model.}
    \label{fig:data-preview}
\end{figure}

High-fidelity DTMs (i.e., 5 cm ground sample distance) of the four prospective Touch-And-Go (TAG) sample sites developed as part of the OSIRIS-REx mission to Asteroid 101955 Bennu, i.e., Nightingale, Kingfisher, Osprey, and Sandpiper, were used to generate ground truth safety map labels for reconnaissance imagery from the mission. 
Specifically, we leverage \textit{monocular} reconnaissance imagery and the corresponding camera pose labels, relative to a body fixed frame of the asteroid, provided through the AstroVision dataset~\cite{driver2022astrovision}. 
For each image, DEMs are constructed by transforming the DTM into a local coordinate system in which the $+z$-axis points opposite the vector corresponding to the direction of the gravitational force due to the target body at the point on the surface closest to the center of the image. 
The gravity due to body was computed using a global shape model of Bennu~\cite{seabrook2022building} and assuming a constant-denity polyhedron~\cite{werner1996exterior}.
Safety mapping was conducted on the DEM and then projected back into the image to produce pixel-wise landing safety labels. 
Example reconnaissance images for each prospective TAG site are provided in Figure \ref{fig:data-preview}. 

\begin{figure}[tb!]
    \centering
    \input{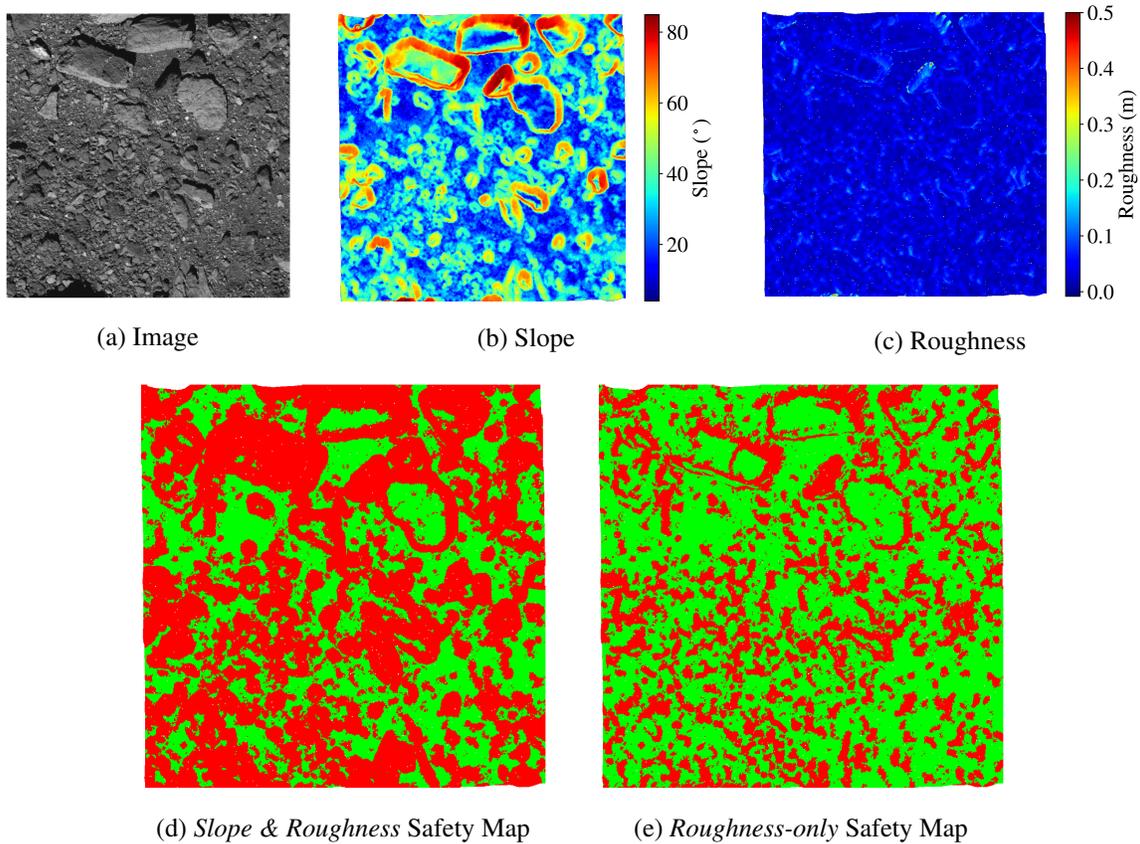}
    \caption{\textbf{Ground truth safety map example.} Safe and unsafe regions are drawn in green and red, respectively, in the safety map.}
    \label{fig:safety-example}
\end{figure}

Landing safety was computed from the DEMs using the method developed by the Autonomous Landing Hazard Avoidance Technology (ALHAT) project~\cite{ivanov2013alhat}. 
The ALHAT method evaluates the lander contact locations for all pixels and for all orientations to assess the worst-case surface slope and roughness values with respect to the surface elevation data contained in the ground truth DEMs. 
Specifically, a landing plane is computed for each pixel by assessing the elevation of four evenly spaced contact points, emulating lander foot pads, on the perimeter of a circle specified by the diameter of the lander. 
Slope is defined as the largest angle between the landing plane and $x$-$y$ plane of the ground truth DEM for all orientations, and the roughness is the largest perpendicular distance to the terrain above the the landing plane for all orientations. 
Any pixel with slope and roughness exceeding a given threshold is labeled as unsafe, where we chose a threshold of 30$^\circ$ for slope and 3.5 cm for roughness. 
We specify a lander with a 35 cm diameter, similar to the MASCOT (Mobile Asteroid surface SCOuT) lander that was deployed during the Hayabusa2 mission to Asteroid 162173 Ryugu~\cite{ho2017mascot}.
An example safety map along with its corresponding monocular image is provided in Figure \ref{fig:safety-example}.

\begin{figure}[tb!]
    \centering
    \includegraphics[width=\linewidth]{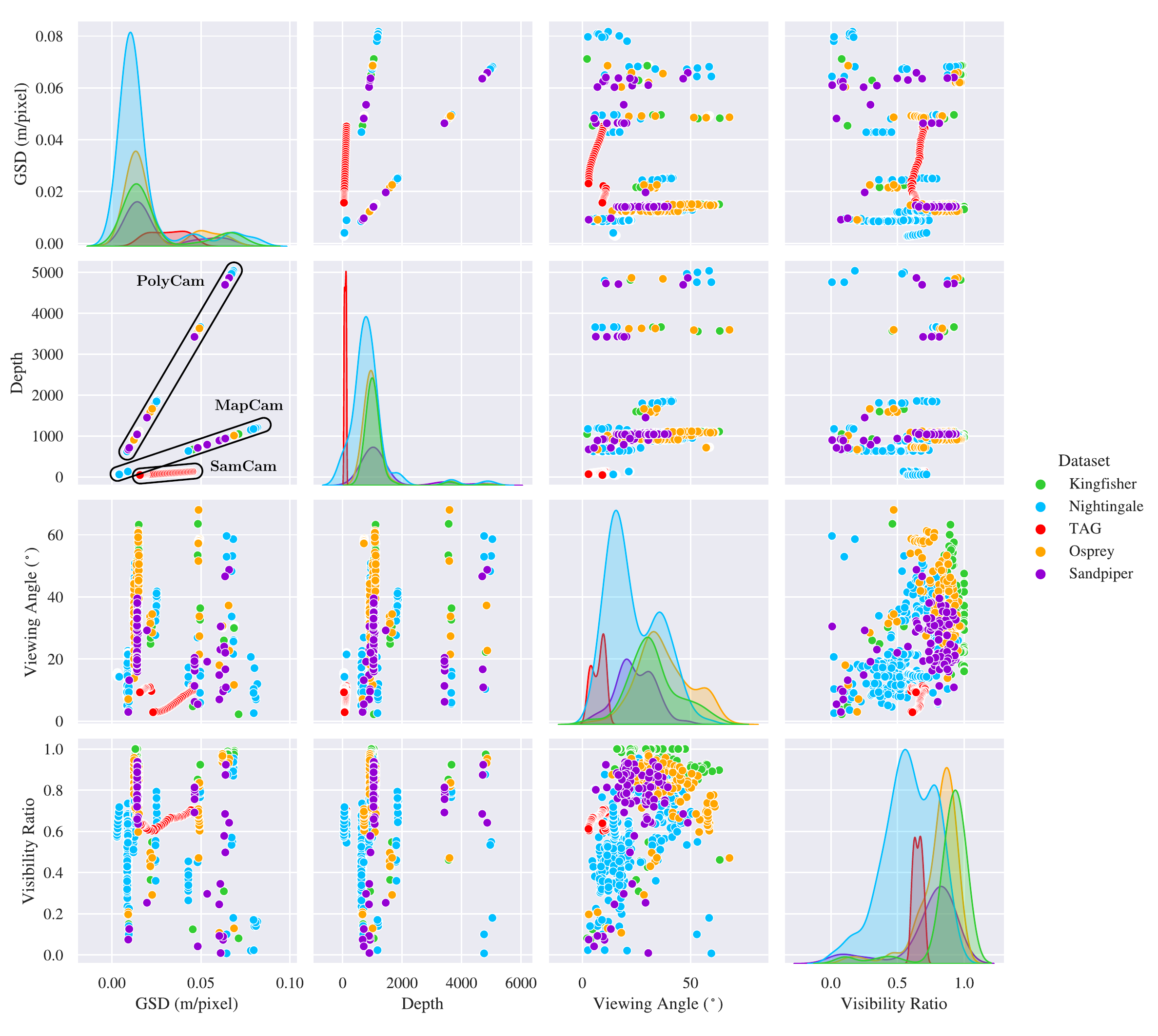}
    \caption{\textbf{Data distributions with respect to imaging depth, GSD, viewing angle, and visibility ratio.} Our dataset features a total of 770 images annotated with per-pixel safety labels: 133 of Kingfisher, 342 of Nightingale, 162 of Osprey, and 91 of Sandpiper, and 42 from the TAG sample collection event at Nightingale.}
    \label{fig:data-distributions}
\end{figure}

Moreover, we provide the data distributions of our datasets with respect to the ground sample distance, imaging depth, viewing angle, and visibility ratio in Figure \ref{fig:data-distributions}. 
\textit{Ground sample distance} (GSD) measures the average distance on the surface spanned by a single pixel, which is a function of the distance to the surface and the camera intrinsics, as landing safety becomes increasingly difficult to observe as the relative size of the lander in the image decreases. 
The \textit{imaging depth} measures the average distance to the surface when the image was taken, and provides context for the GSD values. 
Specifically, the MapCam of the OSIRIS-REx Camera Suite (OCAMS)~\cite{rizk2018ocams}, with a focal length of $\sim$125 mm, can provide 5 cm GSD measuresments of the surface at distances of approximately 1 km, while the PolyCam, with a focal length of $\sim$620 mm, provides the same resolution at distances of almost 4 km.
\textit{Viewing angle} measures the angle between the $-z$-axis of the ground truth DEM and the camera boresight. 
Finally, the \textit{visibility ratio} is the ratio of visible (i.e., not occluded by shadows) pixels to total pixels in the image and provides a measure of the illumination conditions in the image.


\section{Results}

In this section, we first present our suite of metrics used to evaluate the performance of our approach. 
We then validate our approach on two different experiments using real images from the OSIRIS-REx mission to Asteroid 101955 Bennu, including images captured during the actual TAG sample collection event. 

\subsection{Metric Definitions} \label{sec:metrics}

We measure the quality of the predicted per-pixel safety map labels of our model with respect to precision, sensitivity, accuracy, and mean intersection over union (mIoU):
\begin{equation}
    \text{precision} = \frac{\text{true safe}}{\text{true safe} + \text{false safe}},
\end{equation}
\begin{equation}
    \text{sensitivity} = \frac{\text{true safe}}{\text{true safe} + \text{false unsafe}},
\end{equation}
\begin{equation}
    \text{accuracy} = \frac{\text{true safe} + \text{true unsafe}}{\text{valid pixels}},
\end{equation}

\begin{equation}
    \text{mIoU} =\frac{1}{2}\left(\frac{\text{true safe}}{\text{valid pixels} - \text{true unsafe}} + \frac{\text{true unsafe}}{\text{valid pixels} - \text{true safe}}\right).
\end{equation}
\textit{True safe} (\textit{false safe}) includes pixels predicted to be safe by our models that are safe (unsafe) in the ground truth labels, and \textit{true unsafe} (\textit{false unsafe}) includes pixels predicted to be unsafe that are unsafe (safe) in the ground truth labels. 
Note that false unsafe includes safe pixels that are ignored and not labeled safe due to high uncertainty. 
For our application, we can interpret \textit{precision} as the reliability of the pixels predicted to be safe, and \textit{sensitivity} as detection rate of true safe sites, respectively. 
\textit{Accuracy} and \textit{mIoU} are the metrics evaluated for the valid pixels, which are the pixels with smaller uncertainty than the threshold. 
In other words, valid pixels correspond to the predictions that the network is most ``certain" about. 
For the results without uncertainty thresholding, accuracy and mIoU are evaluated for all the pixels with valid safety labels. 
In the following analysis, we refer to the ratio of pixels that fall above our uncertainty threshold, and consequently marked as unsafe, to valid pixels as the \textit{screening rate}.


\subsection{Experiment 1: Prospective Landing Site Sandpiper}

\begin{table*}[tp!]
\footnotesize
\centering
\scshape
\ra{1.5}
\caption{\textbf{Overall performance for the Sandpiper landing site experiment.} The values in parentheses are the metrics with shadowed pixels ignored. All reported values are percentages.}
\begin{adjustbox}{width=\linewidth}
\begin{tabular}{lrrrr}
\toprule
Method & Precision & Sensitivity & Accuracy & mIoU \\
\midrule 
\multicolumn{5}{l}{\textbf{Slope \& roughness}} \\
Without uncertainty   & 60.66 (62.91) & \textbf{67.21} (\textbf{70.05}) & 69.53 (69.41) & 52.05 (52.27) \\
With uncertainty      & \textbf{76.98} (\textbf{77.67}) & 20.09 (21.86) & \textbf{82.29} (\textbf{82.01}) & \textbf{65.76} (\textbf{65.93}) \\
\midrule 
\rowcolor[gray]{0.9}
\multicolumn{5}{l}{\textbf{Roughness only}} \\
\rowcolor[gray]{0.9}
Without uncertainty   & 77.24 (78.92) & \textbf{61.61} (\textbf{63.66}) & 63.53 (64.41) & 44.87 (45.31) \\
\rowcolor[gray]{0.9}
With uncertainty      & \textbf{85.71} (\textbf{86.32}) & 28.78 (31.63) & \textbf{73.77} (\textbf{73.93}) & \textbf{55.11} (\textbf{54.98}) \\
\bottomrule
\end{tabular}
\end{adjustbox}
\label{tab:safety-metrics-all-sandpiper}
\end{table*}

\begin{figure}[htbp!]
    \begin{minipage}{0.90\linewidth}
\begin{center}
\begin{subfigure}[t]{\linewidth}
\begin{tabular}{c@{\hskip 5pt}c@{\hskip 5pt}c@{\hskip 0pt}c}
    \includegraphics[height=34mm]{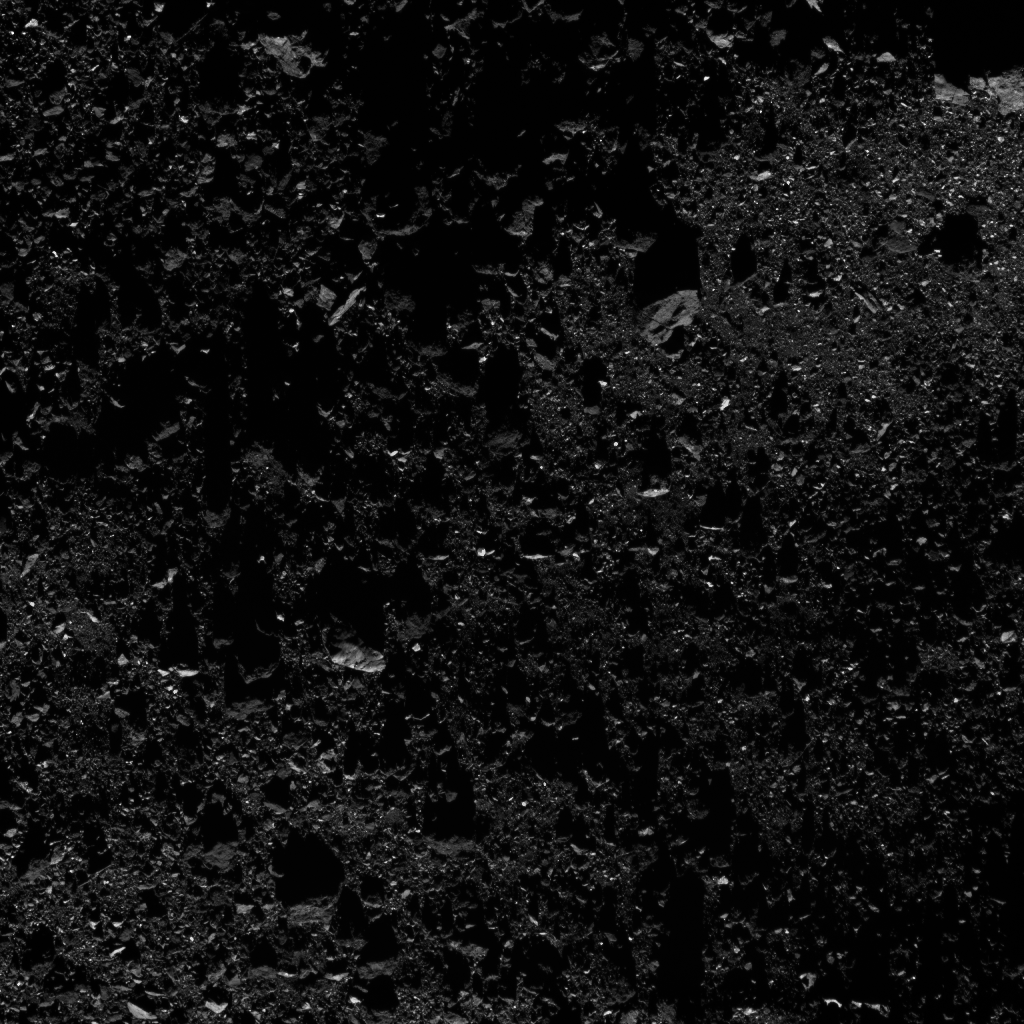} &
    \includegraphics[height=34mm]{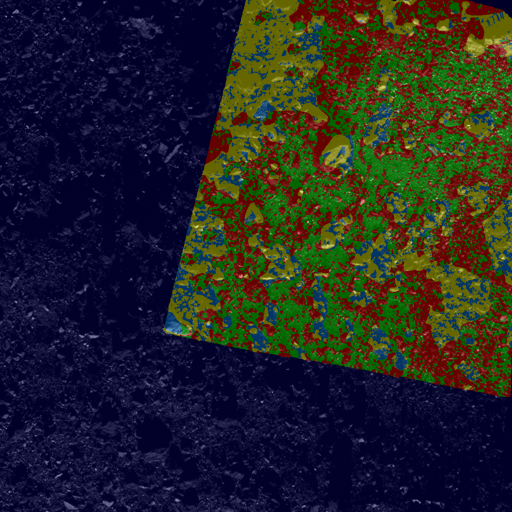} &
    \includegraphics[height=34mm]{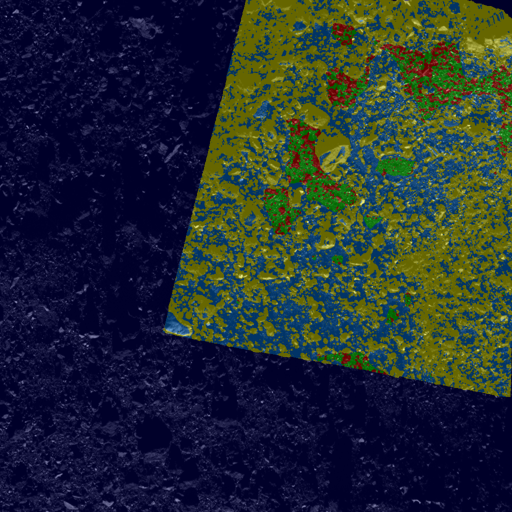} &
    \includegraphics[height=34mm]{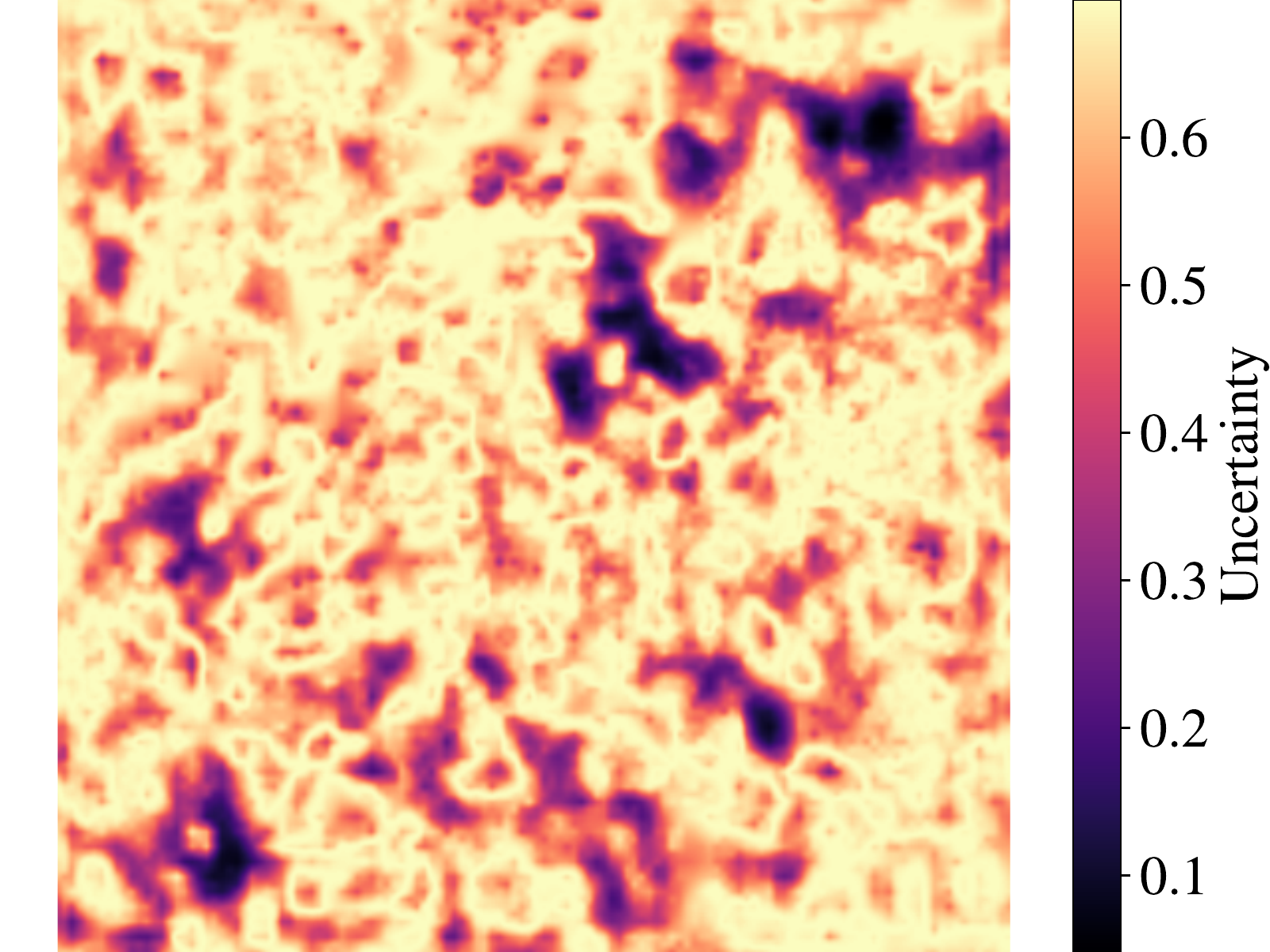} \\
    \includegraphics[height=34mm]{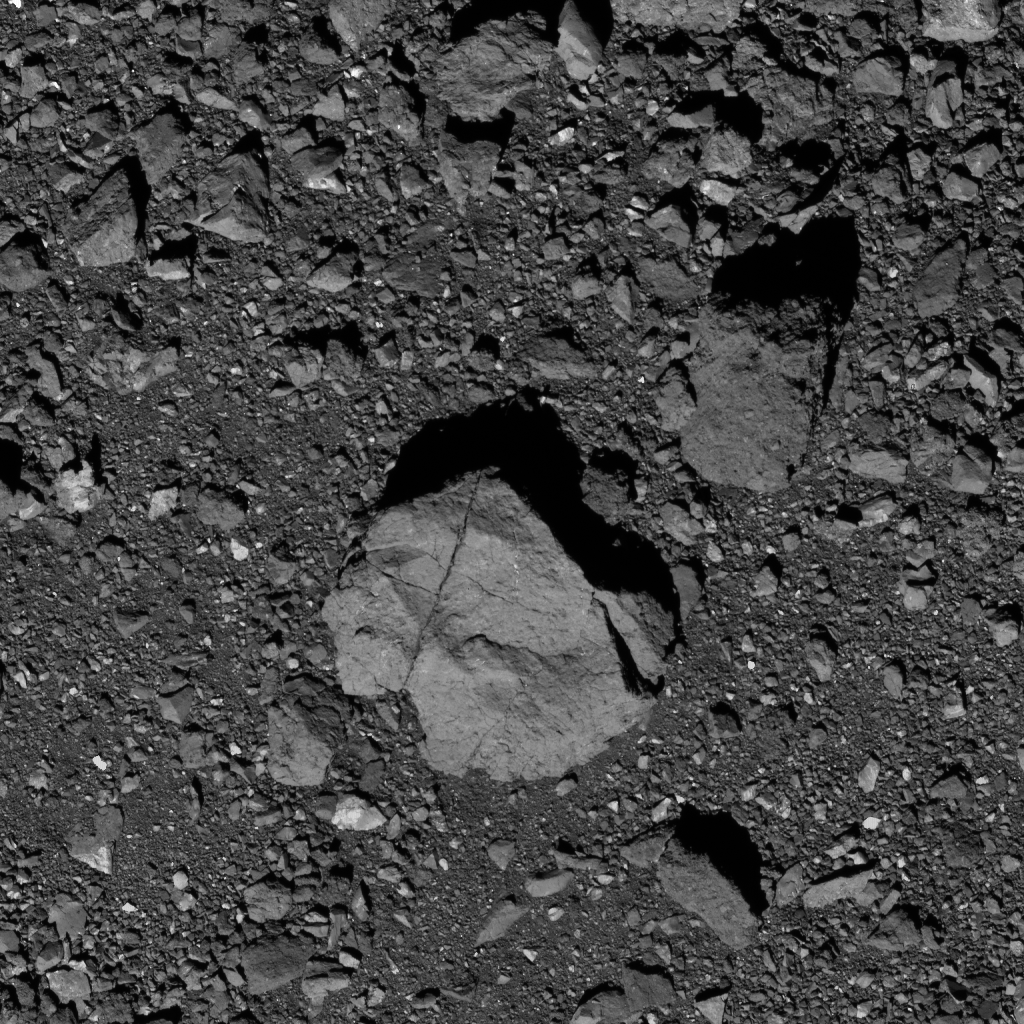} &
    \includegraphics[height=34mm]{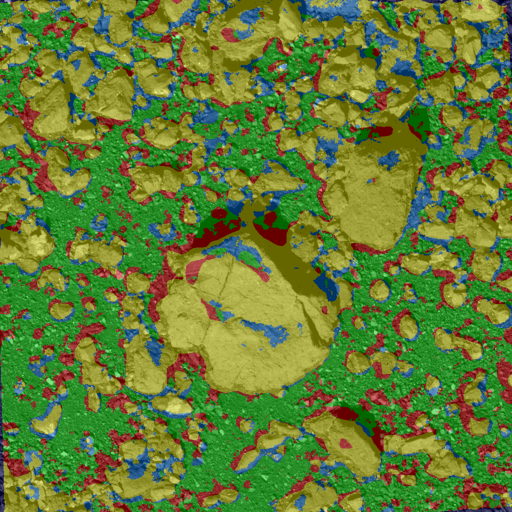} &
    \includegraphics[height=34mm]{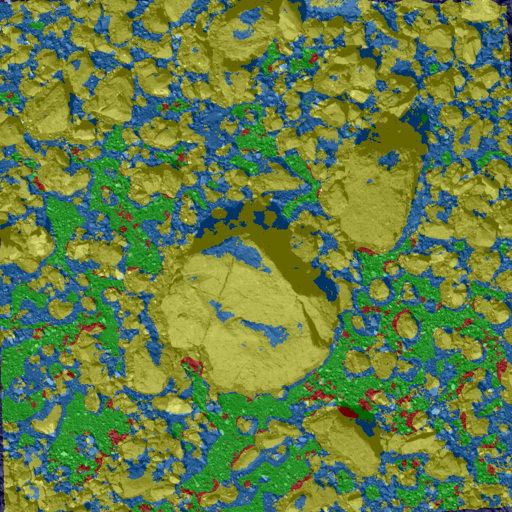} &
    \includegraphics[height=34mm]{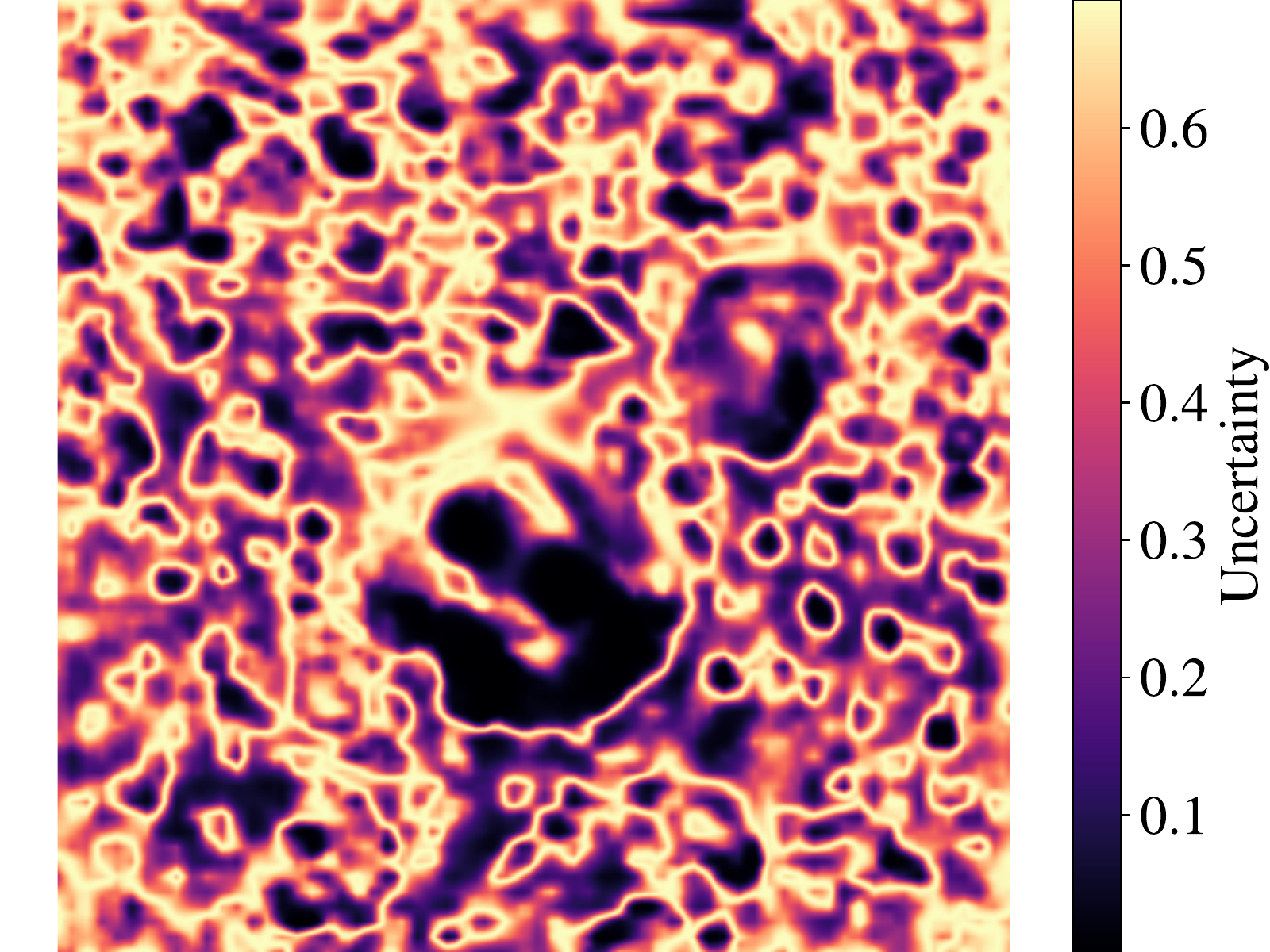} \\
    \includegraphics[height=34mm]{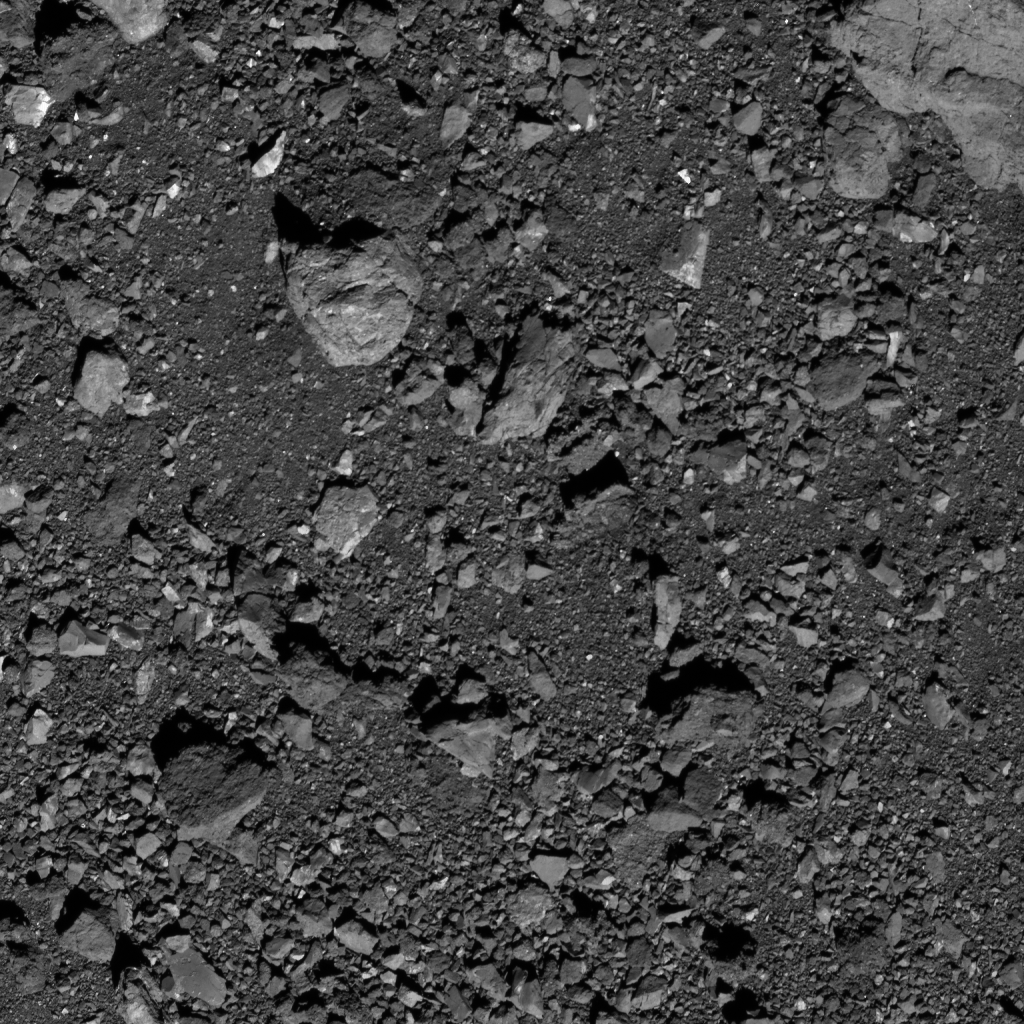} &
    \includegraphics[height=34mm]{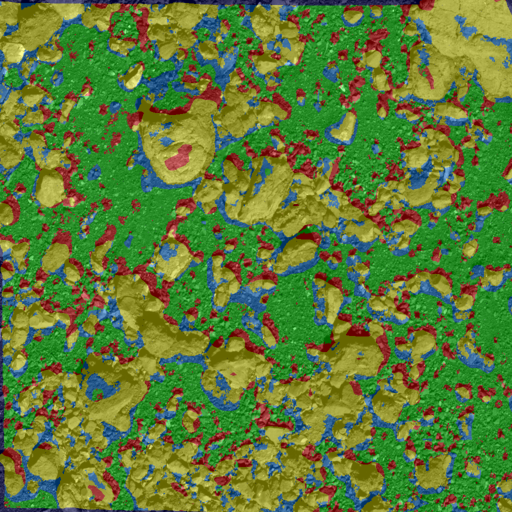} &
    \includegraphics[height=34mm]{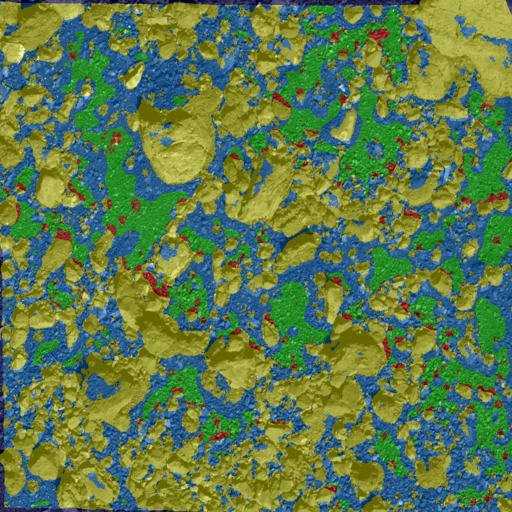} &
    \includegraphics[height=34mm]{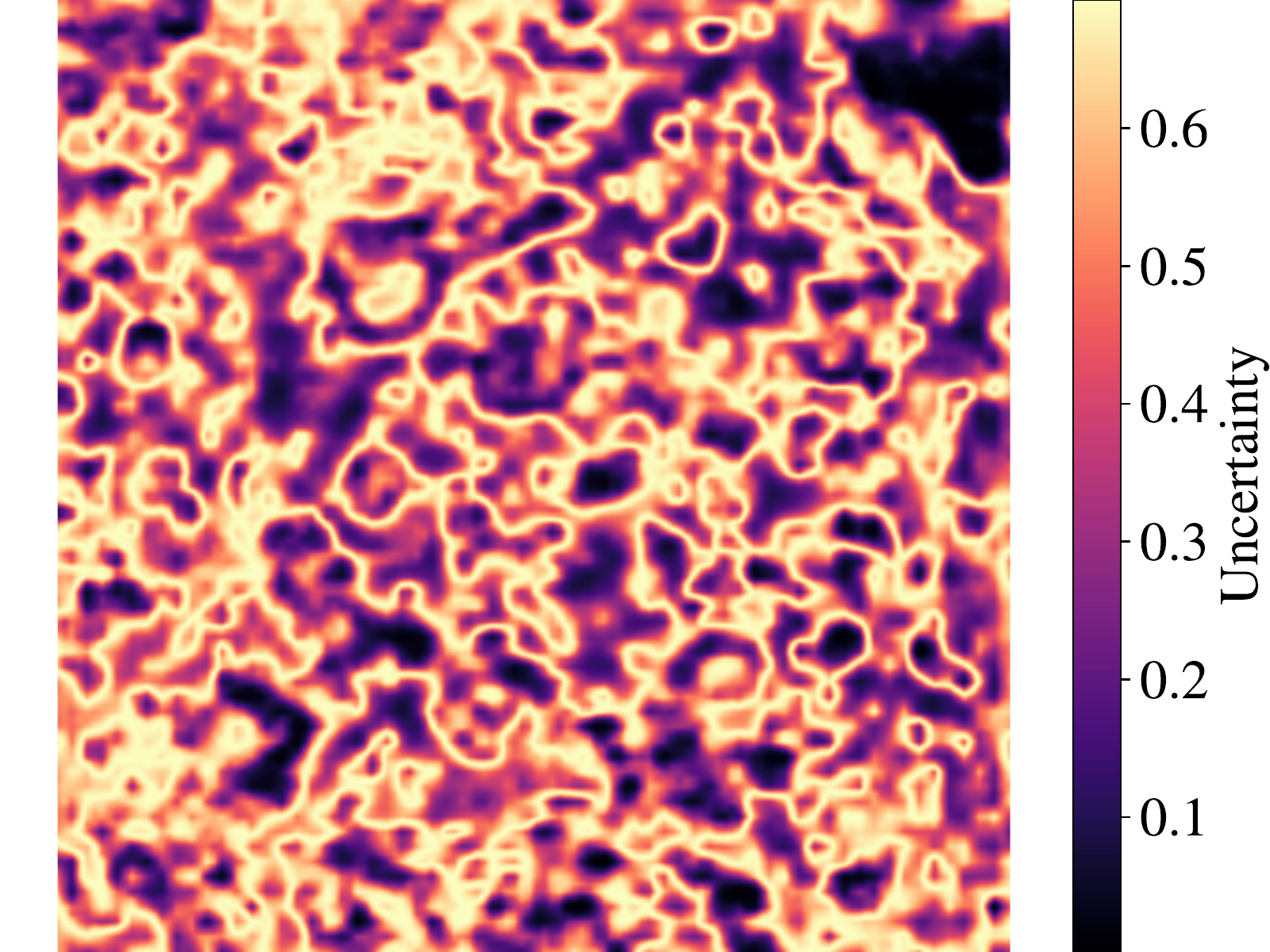} \\
\end{tabular}
\vspace{-5pt}
\caption{\textit{Slope \& roughness}}
\begin{tabular}{c@{\hskip 5pt}c@{\hskip 5pt}c@{\hskip 0pt}c}
    \includegraphics[width=34mm]{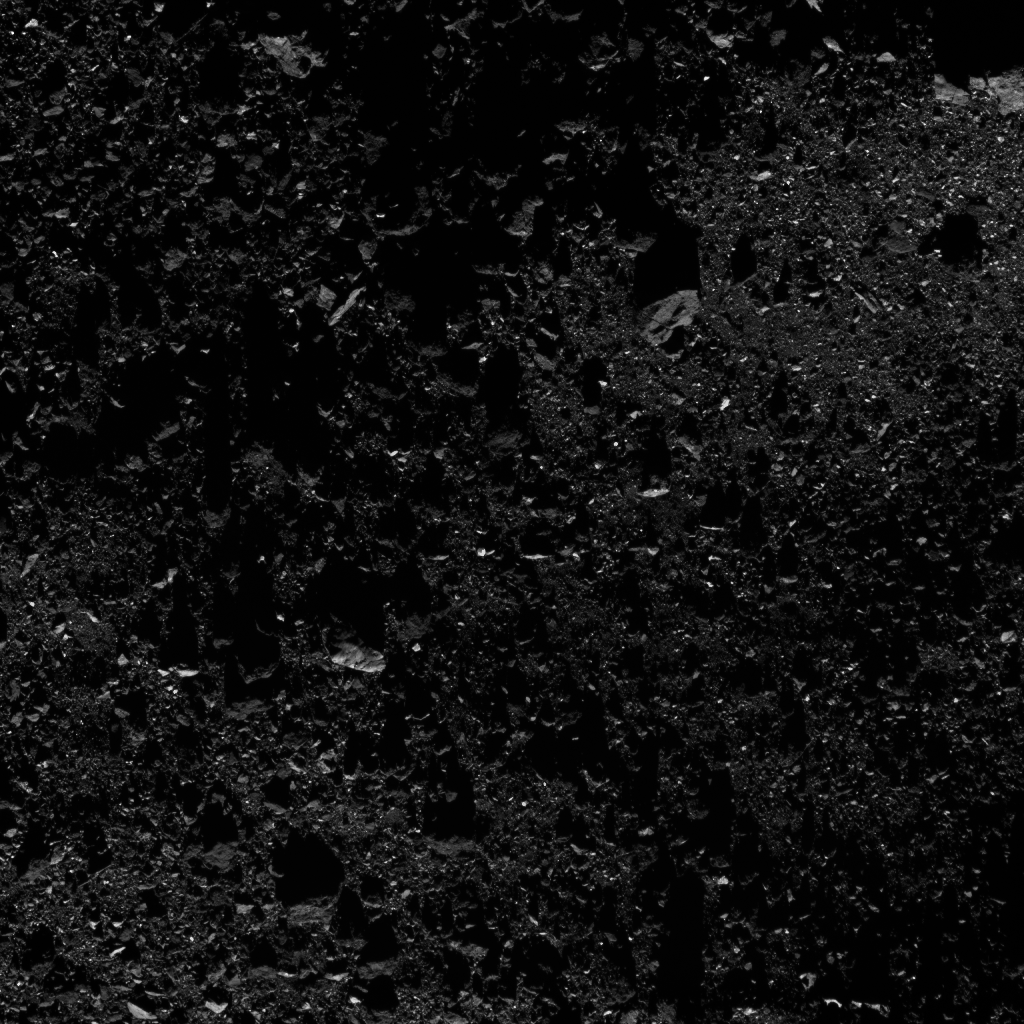} &
    \includegraphics[height=34mm]{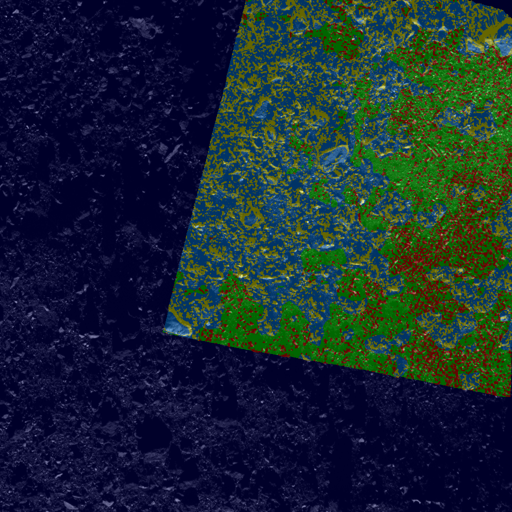} &
    \includegraphics[height=34mm]{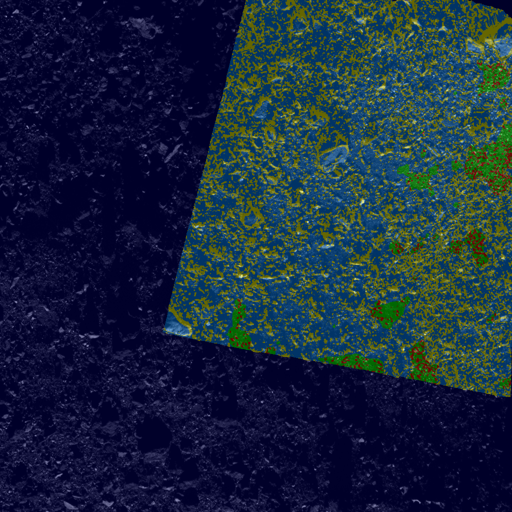} &
    \includegraphics[height=34mm]{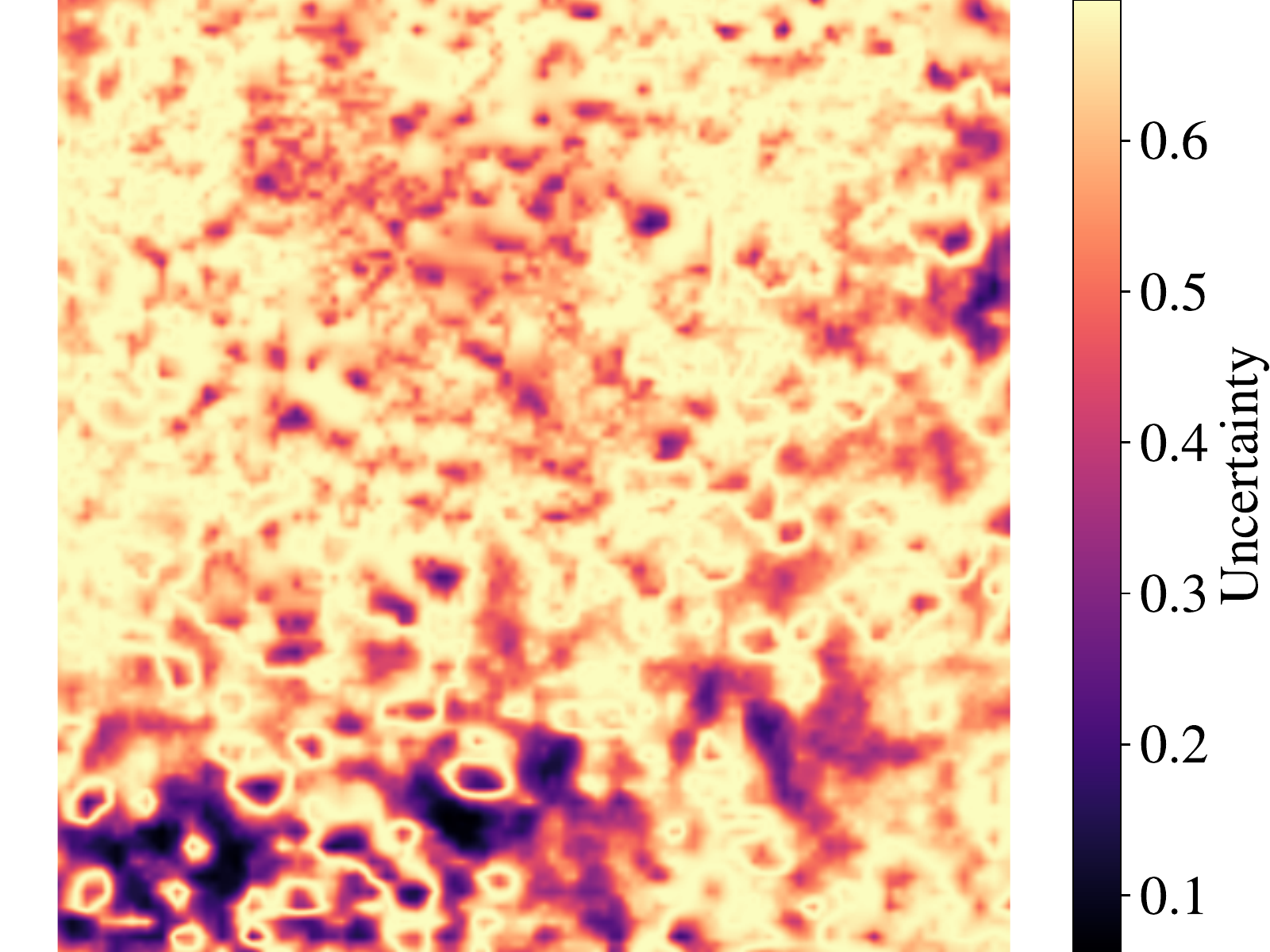} \\
    \includegraphics[height=34mm]{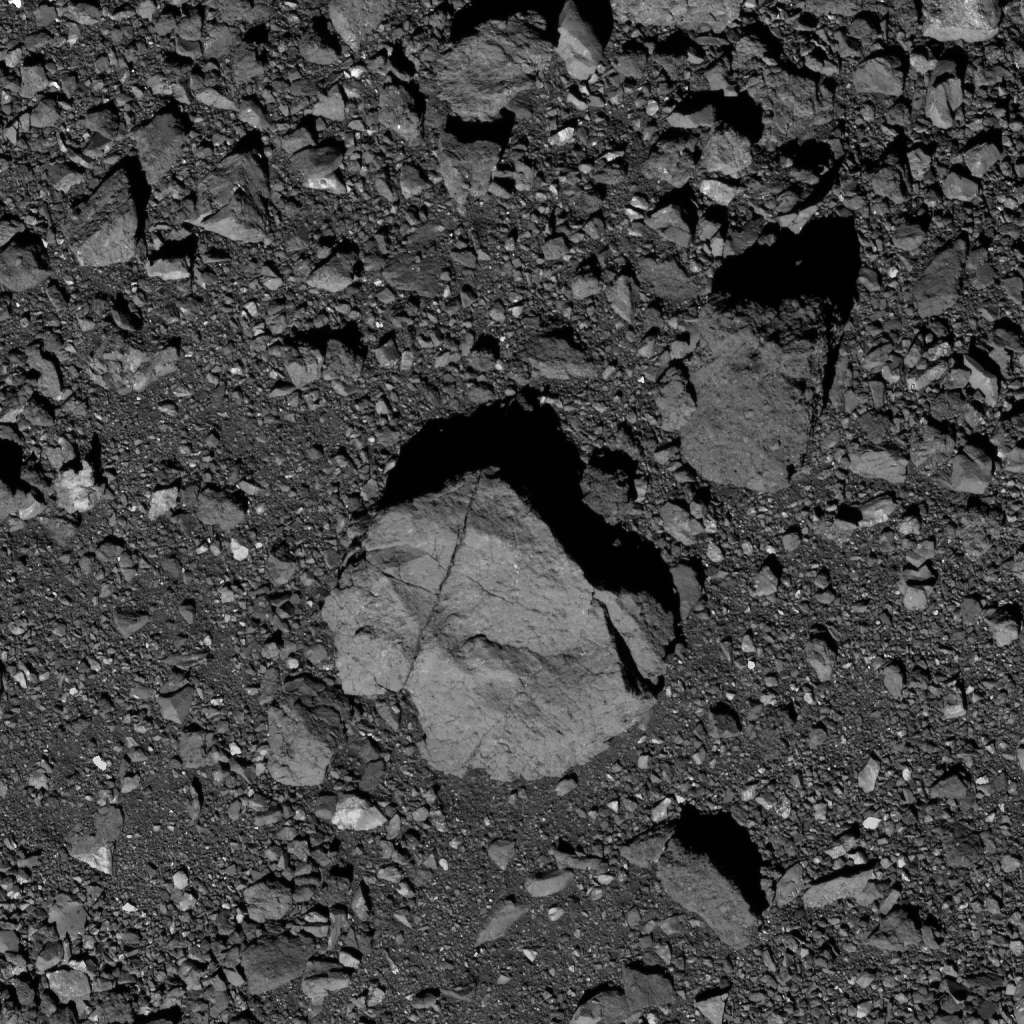} &
    \includegraphics[height=34mm]{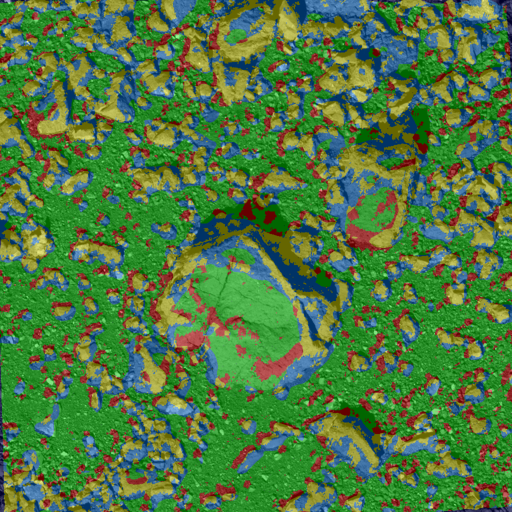} &
    \includegraphics[height=34mm]{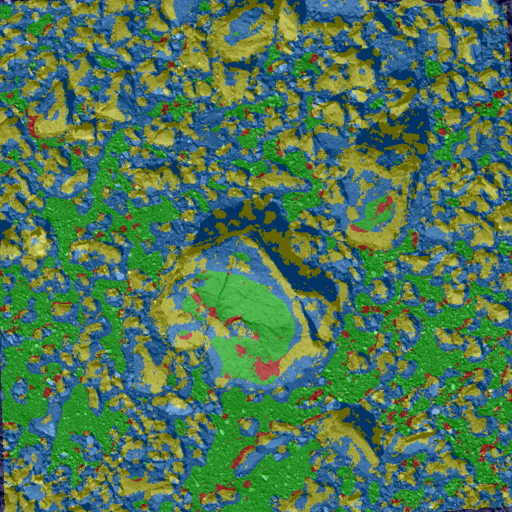} &
    \includegraphics[height=34mm]{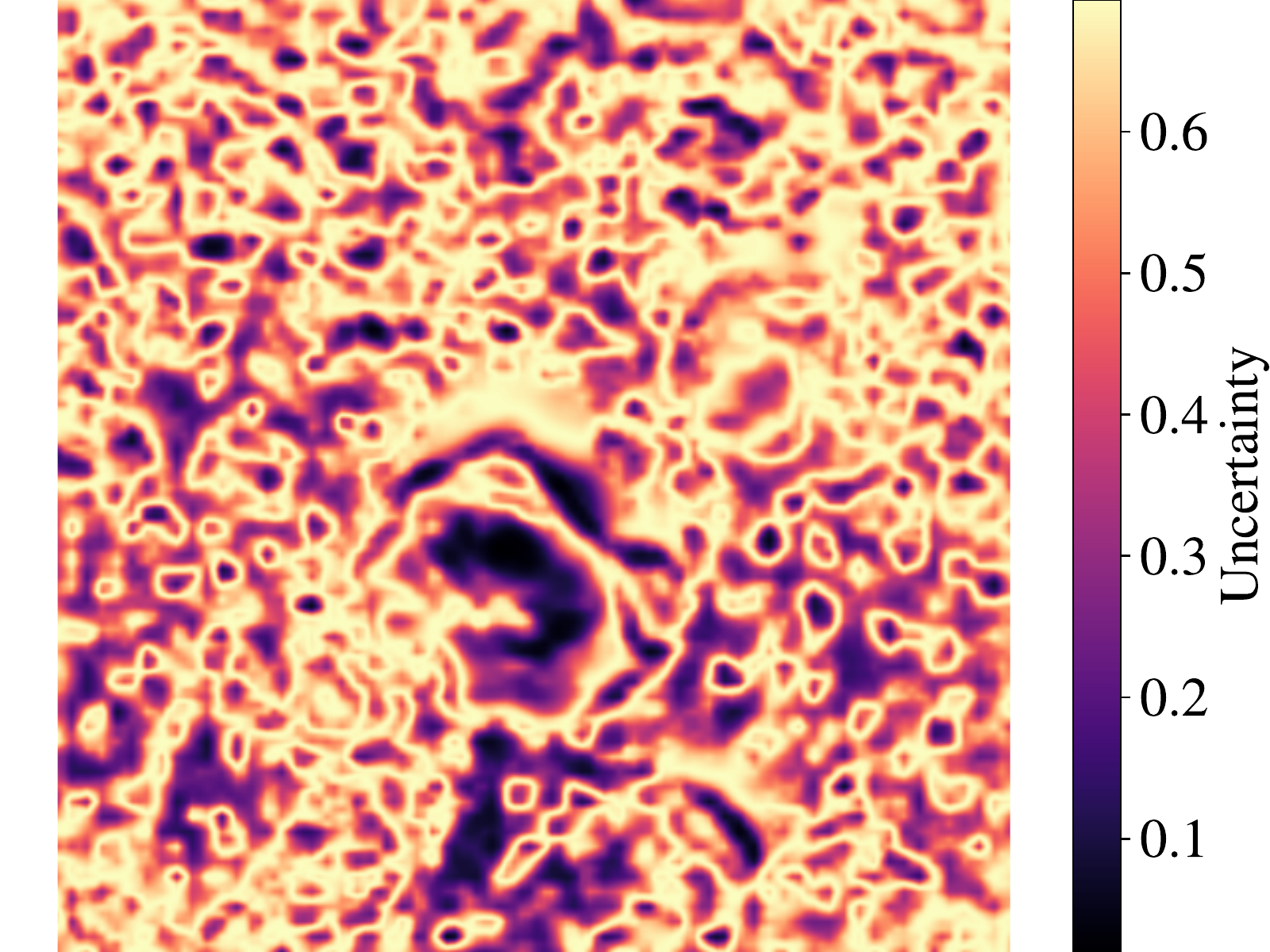} \\
    \includegraphics[height=34mm]{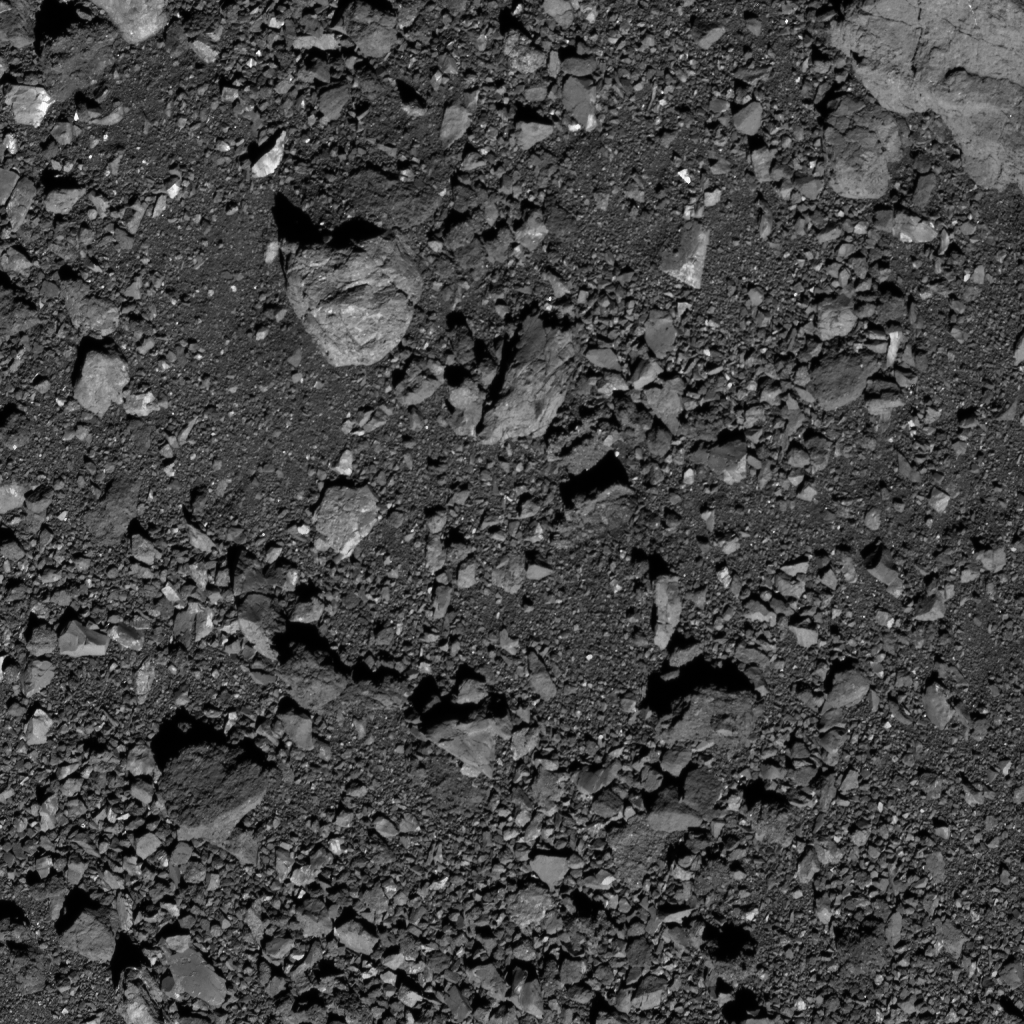} &
    \includegraphics[height=34mm]{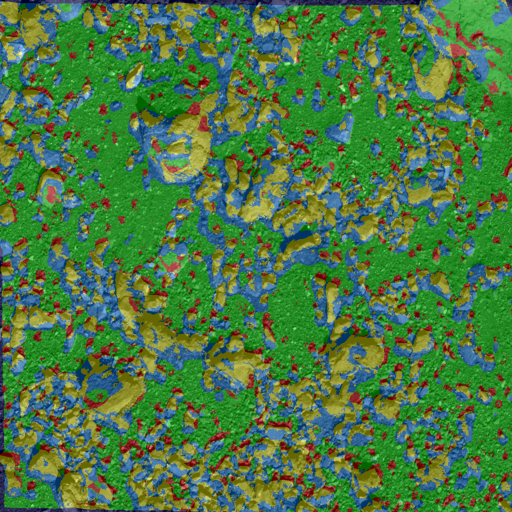} &
    \includegraphics[height=34mm]{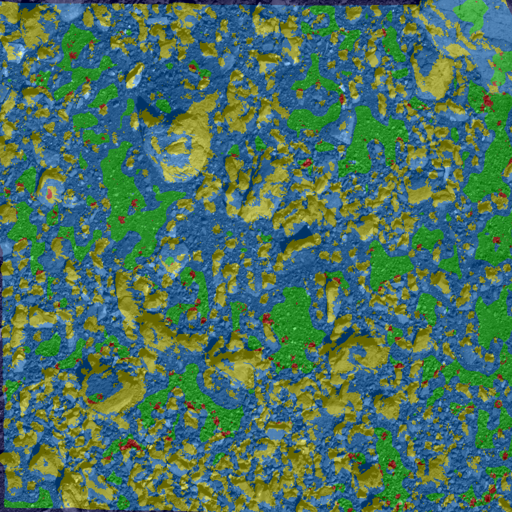} &
    \includegraphics[height=34mm]{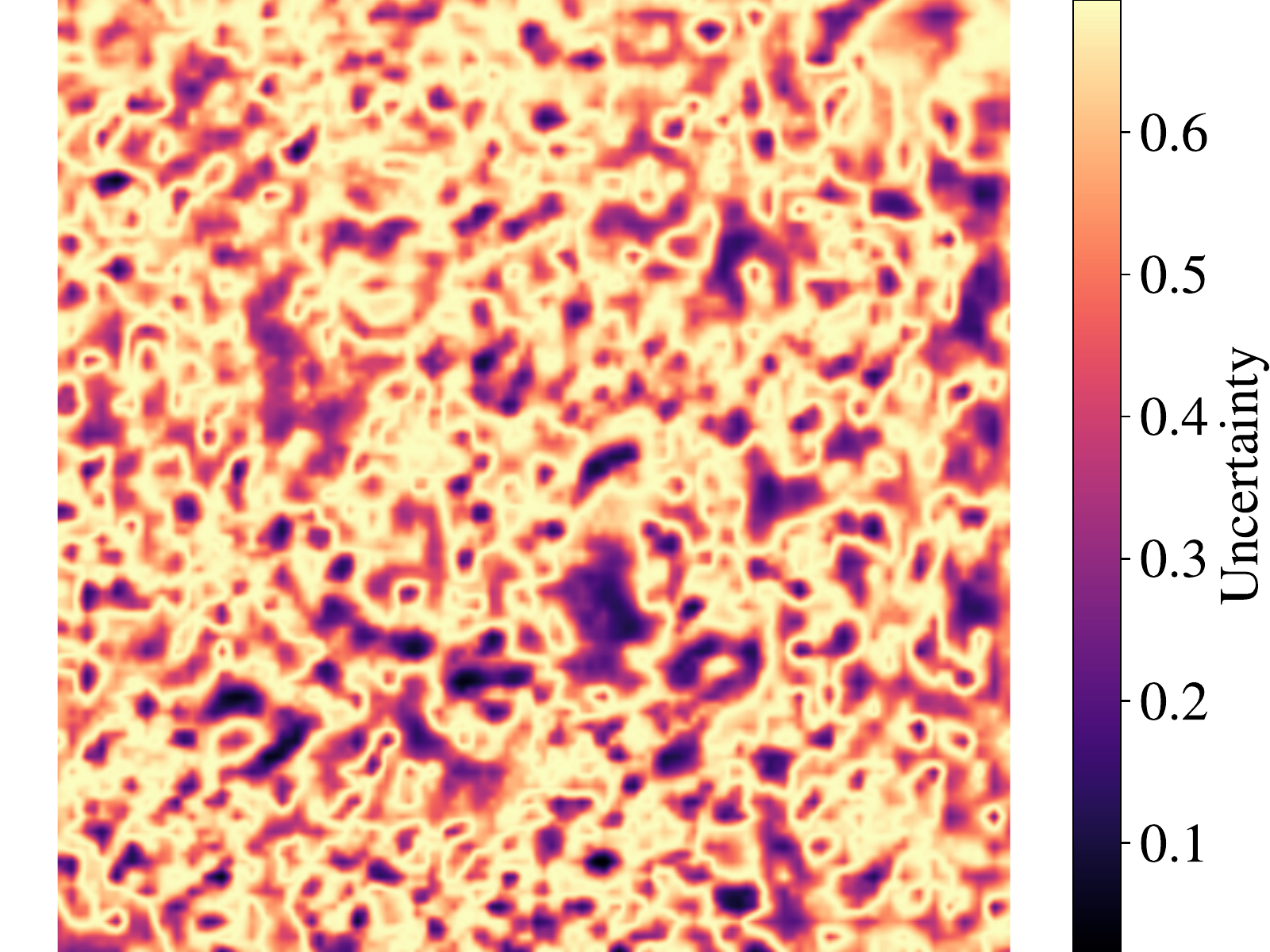} \\
    \footnotesize{Test Image} & \footnotesize{Without Uncertainty} & \footnotesize{With Uncertainty} & \footnotesize{Uncertainty}\\
\end{tabular}
\vspace{-5pt}
\caption{\textit{Roughness-only}}
\end{subfigure}\\
\end{center}
\end{minipage}
    \vspace{-5pt}
    \caption{\textbf{Qualitative monocular safety mapping results for the Sandpiper experiment.} Green, yellow, blue, and red labels represent true safe, true unsafe, false unsafe, and false safe, respectively.}
    \label{fig:qual-sandpiper-safety}
\end{figure}

We train our model on images from three of the prospective landing sites from the OSIRIS-Rex mission, namely, Nightingale, Osprey, and Kingfisher, and test our model on images of the remaining sample site, Sandpiper. 
This emulates a scenario in which data from previously mapped landing sites could be used to train a network to predict landing safety in a new, unexplored region of the target body without requiring the construction of high-fidelity DEMs. 
These results are detailed in Table \ref{tab:safety-metrics-all-sandpiper}, and qualitative examples are provided in Figure \ref{fig:qual-sandpiper-safety}, where we consider performance with respect to identifying both slope and roughness hazards and roughness-only hazards. 

The results illustrate that our models are able to predict safety maps from just a single monocular image of the propspective landing site, which is completely unseen during training, with accuracy over $69\%$ for the slope and roughness hazard detection case, and over $63\%$ for the roughness-only hazard detection case, even without uncertainty thresholding. 
Moreover, our Bayesian ICNet architecture enables uncertainty thresholding in order to further boost performance by ignoring regions in which the models' prediction has high entropy. 
With the uncertainty threshold, accuracy increases to $82.29\%$ and $73.77\%$ for the slope and roughness and roughness-only cases, respectively, at the cost of decreased sensitivity. 
Importantly, we are able to achieve $76.98\%$ and $85.71\%$ precision for the slope and roughness and roughness-only cases, respectively, after uncertainty thresholding.
We also have a slight increase in all the metrics by ignoring shadowed pixels, which are reported in parentheses in Table \ref{tab:safety-metrics-all-sandpiper}.

Comparing the two different hazard detection tasks, i.e., slope and roughness hazards and roughness-only hazards, the roughness-only case has lower values of sensitivity, accuracy, and mIoU, but higher values of precision. 
This suggests that the roughness-only case is a harder task in terms of precisely labeling safe and unsafe pixels on average, resulting in lower accuracy and mIoU, but is an easier task in terms of identifying only safe pixels, thus resulting in higher precision. 
This is partially due to the higher incidence of safe pixels for the roughness-only case as compared to the slope and roughness case, as illustrated in Figure \ref{fig:safety-example}. 

\begin{figure}
    \centering
    \begin{subfigure}[c]{0.72\linewidth}
        \includegraphics[width=\linewidth]{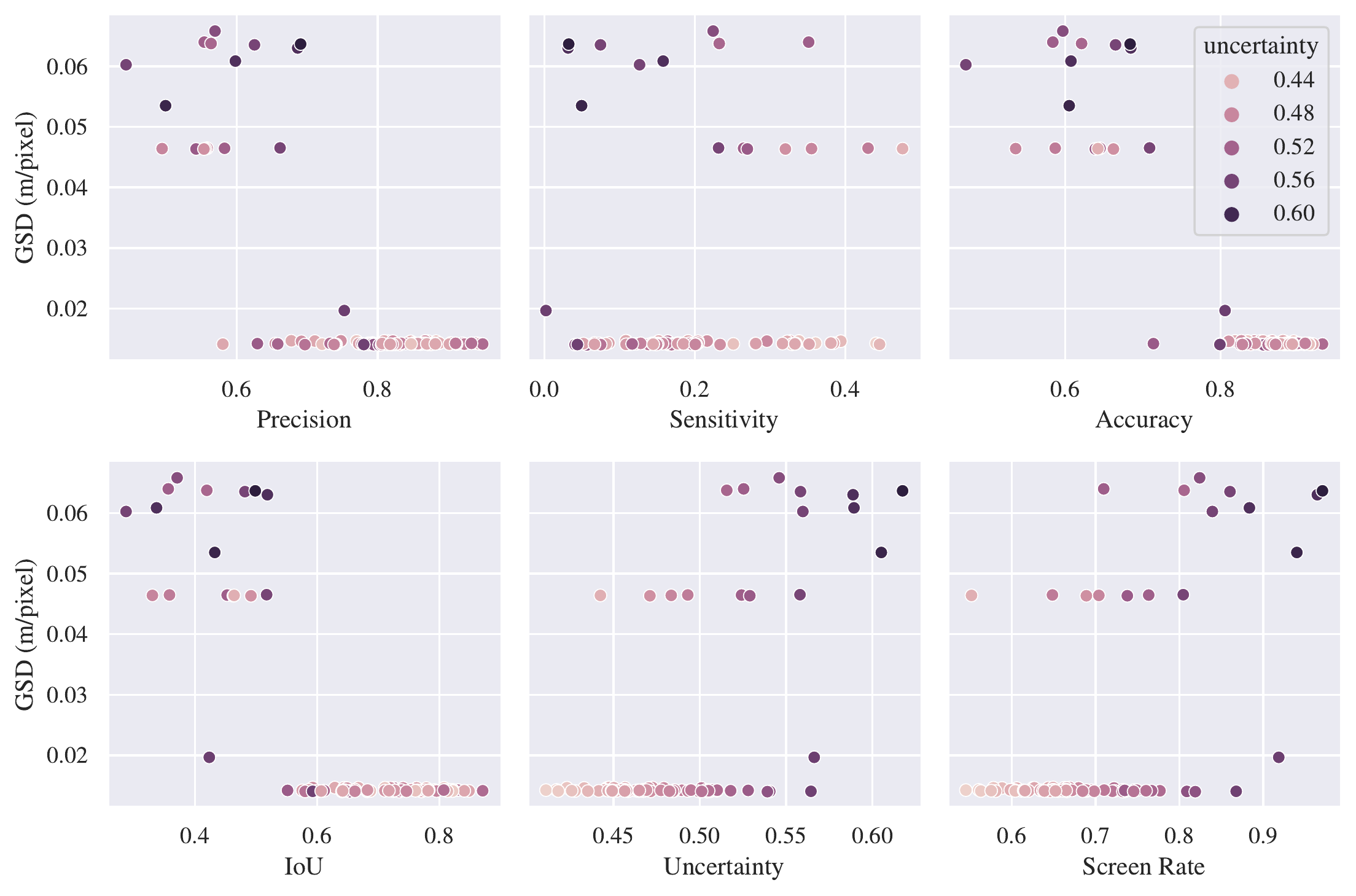}
    \end{subfigure}\\
    \vspace{12pt}
    \begin{subfigure}[c]{0.72\linewidth}
        \includegraphics[width=\linewidth]{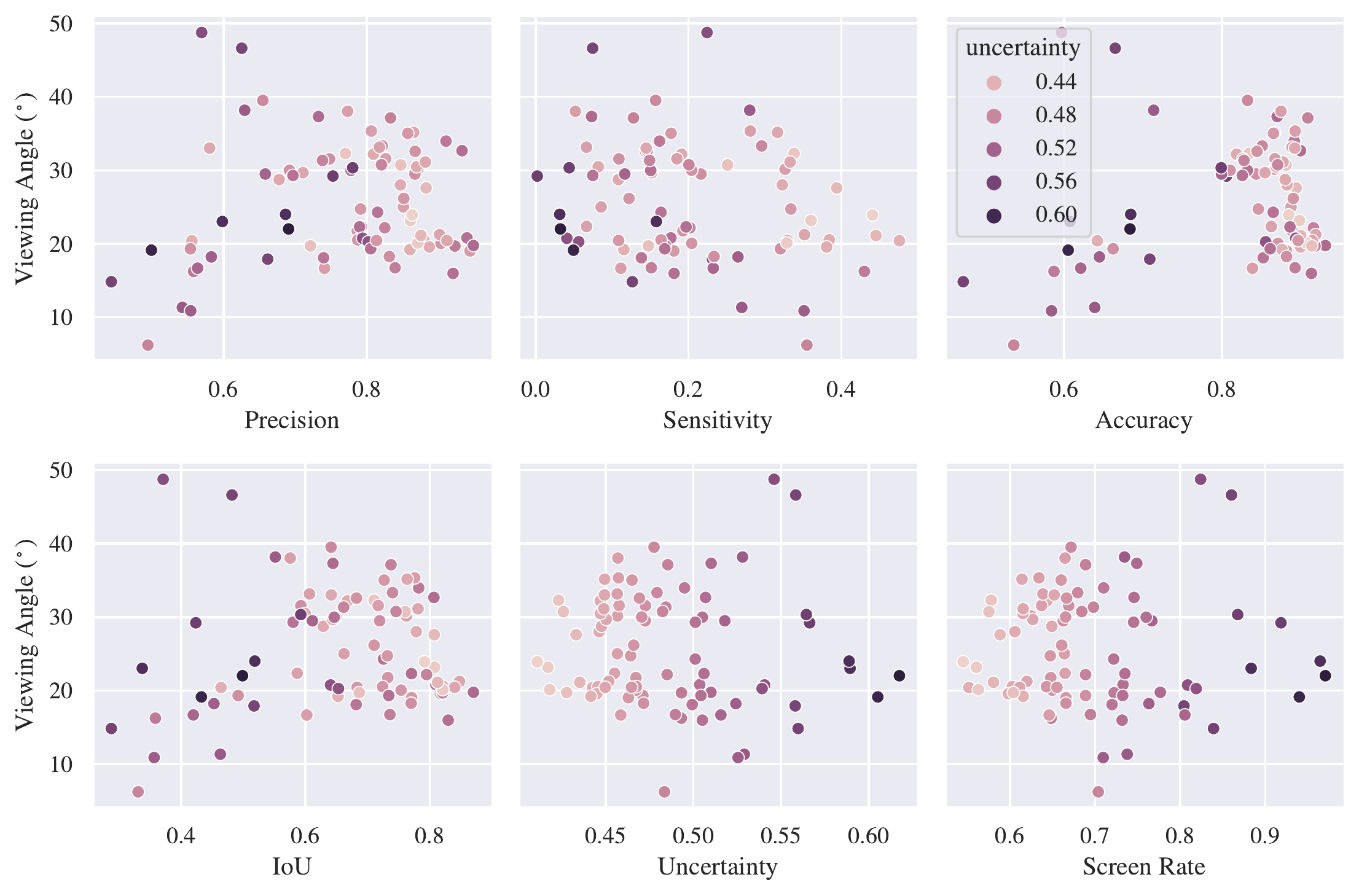}
    \end{subfigure}\\
    \vspace{12pt}
    \begin{subfigure}[c]{0.72\linewidth}
        \includegraphics[width=\linewidth]{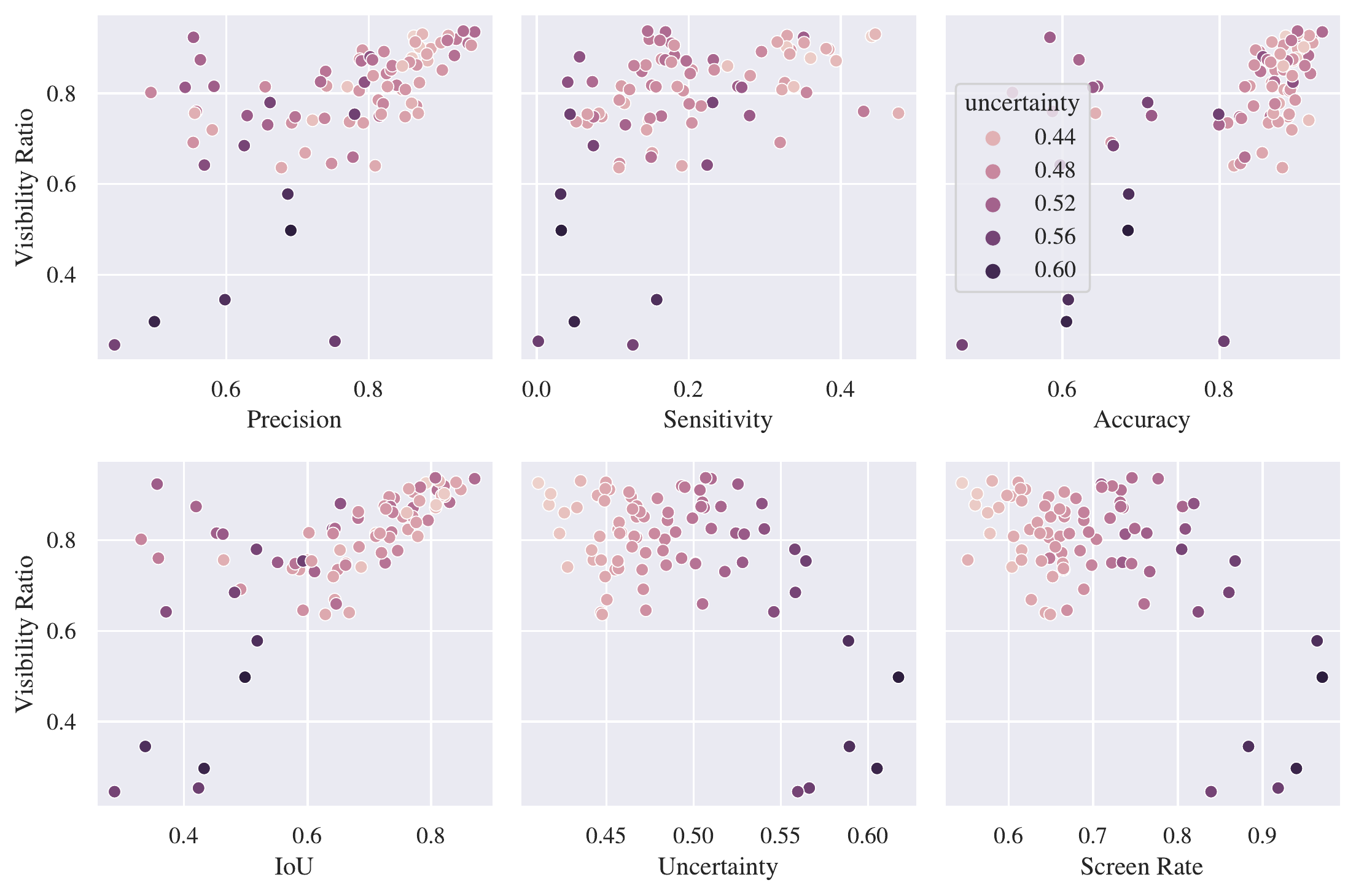}
    \end{subfigure}\\
    \caption{\textbf{Per-image metrics for \textit{slope \& roughness} safety on the Sandpiper experiment with respect to GSD, viewing angle, and visibility ratio.} }
    \label{fig:eval-sandpiper-slope-rghns-all}
\end{figure}

\begin{figure}
    \centering
    \begin{subfigure}[c]{0.72\linewidth}
        \includegraphics[width=\linewidth]{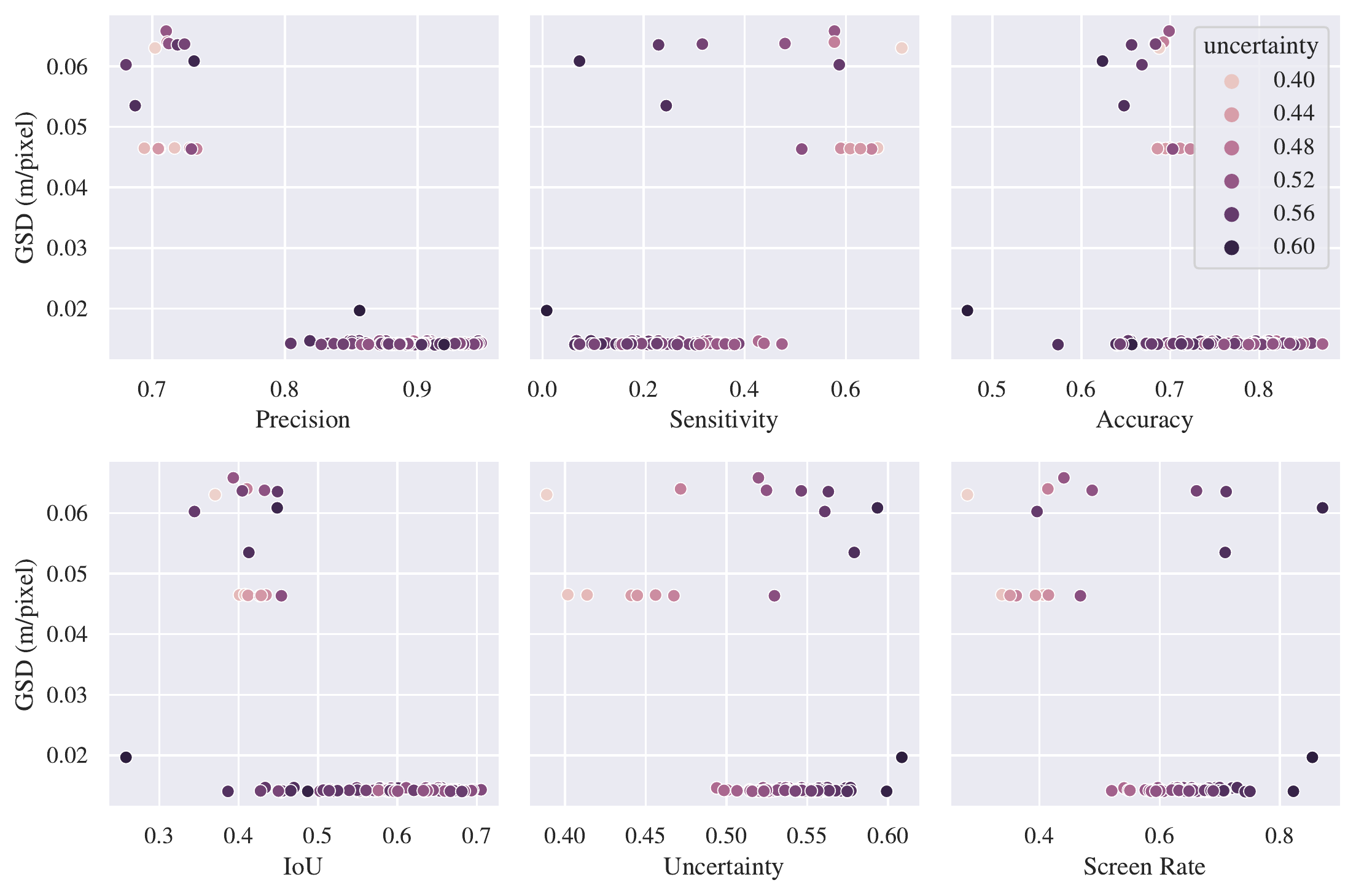}
    \end{subfigure}\\
    \vspace{12pt}
    \begin{subfigure}[c]{0.72\linewidth}
        \includegraphics[width=\linewidth]{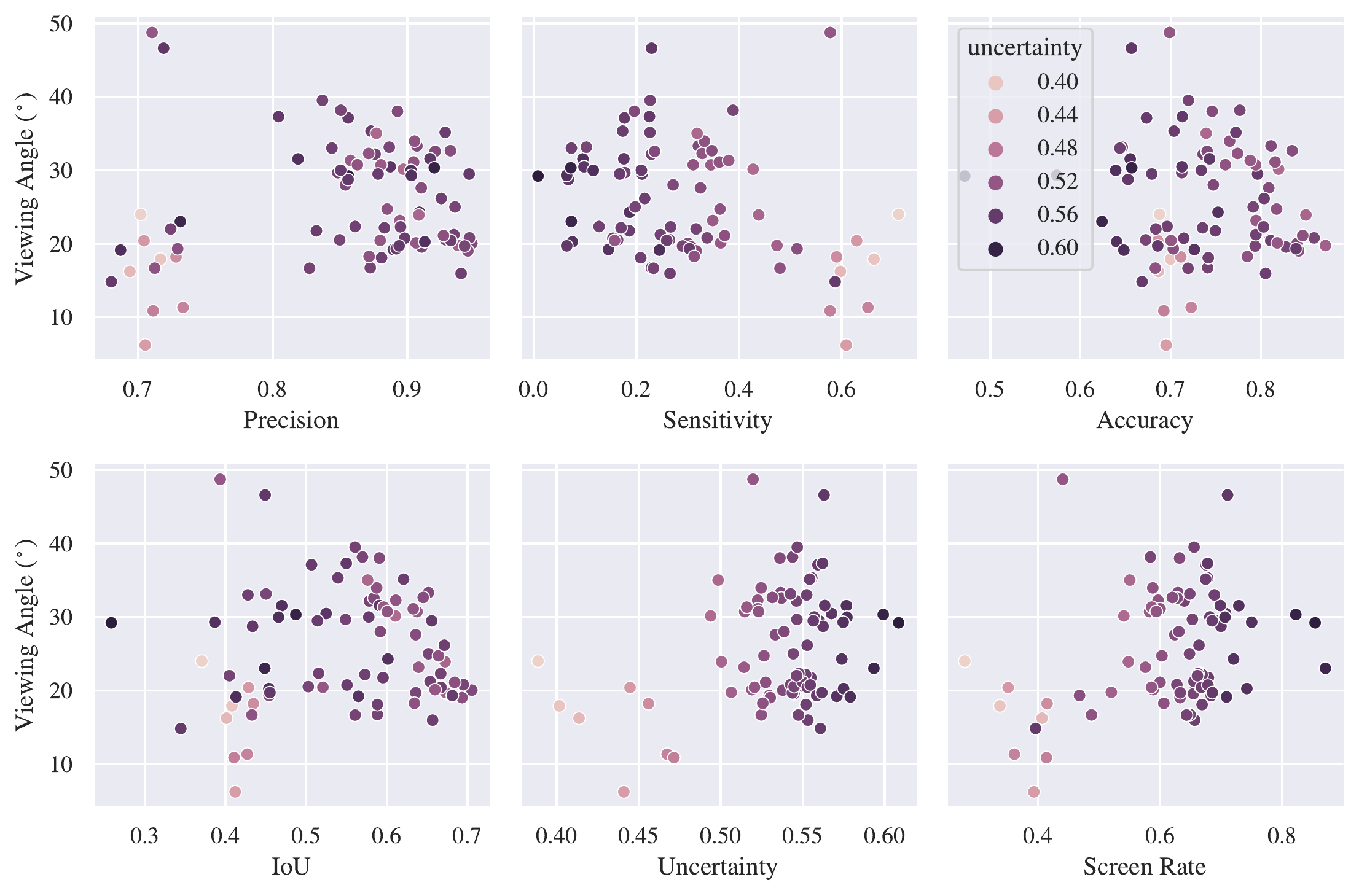}
    \end{subfigure}\\
    \vspace{12pt}
    \begin{subfigure}[c]{0.72\linewidth}
        \includegraphics[width=\linewidth]{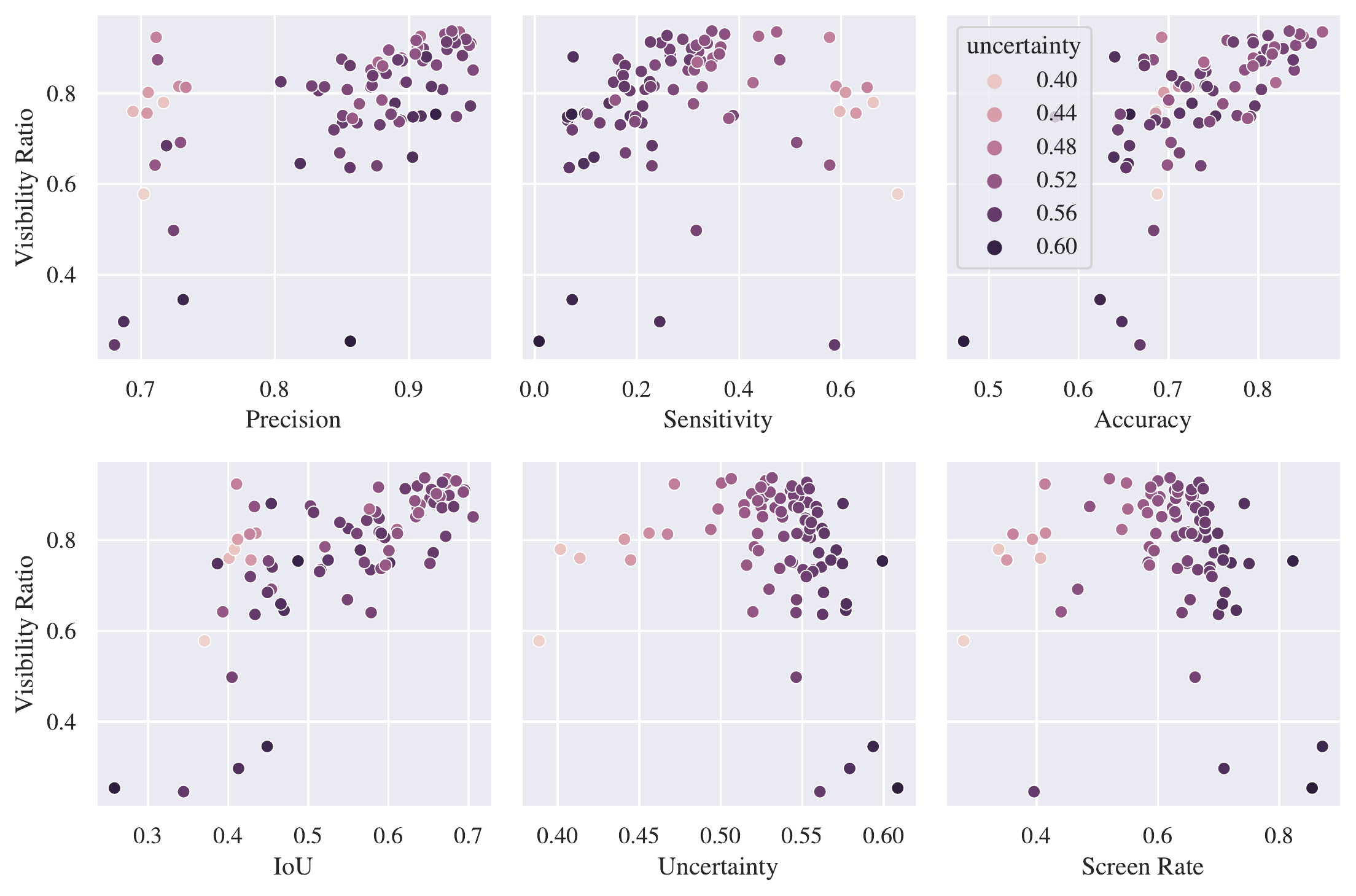}
    \end{subfigure}\\
    \caption{\textbf{Per-image metrics for \textit{roughness-only} safety on the Sandpiper experiment with respect to GSD, viewing angle, and visibility ratio.}}
    \label{fig:eval-sandpiper-rghns-all}
\end{figure}

Additionally, we analyzed the per-image metrics with respect to GSD, viewing angle, and visibility ratio for the slope and roughness case, shown in Figure \ref{fig:eval-sandpiper-slope-rghns-all}, and the roughness-only case, shown in Figure \ref{fig:eval-sandpiper-rghns-all}, in order to identify possible causes of uncertainty in the predictions. 
As a general trend, we can observe that a higher uncertainty results in lower performance metrics of precision, sensitivity, accuracy, and mIoU. 
Intuitively, low visibility is a common factor that results in higher uncertainty in our model for both the slope and roughness case and the roughness-only case. 
Our models assign a higher uncertainty to images with larger GSD for the slope and roughness case, and larger viewing angle for the roughness-only case. 
Note that increased uncertainty for images at higher GSDs may also be due to these instances being less represented in the training data as shown in Figure \ref{fig:data-distributions}. 
In either case, we demonstrate that the uncertainty threshold serves as a powerful tool for detecting and accounting for difficult or out-of-distribution input conditions, allowing our models to predict precise and accurate safety maps across multiple GSDs, viewing angles, and illumination conditions.

\subsection{Experiment 2: OSIRIS-REx TAG Sequence}

\begin{figure}[htbp!]
    \begin{minipage}{\linewidth}
\begin{center}
\begin{tabular}[b]{@{}c@{\hskip 5pt}c@{\hskip 5pt}c@{\hskip 5pt}c@{\hskip 5pt}c@{\hskip 5pt}c@{\hskip 5pt}c@{\hskip 5pt}c@{}}
    \texttt{\tiny{2020-10-20T21:30:48}} & \texttt{\tiny{2020-10-20T21:31:48}} & \texttt{\tiny{2020-10-20T21:32:48}} & \texttt{\tiny{2020-10-20T21:33:48}} & \texttt{\tiny{2020-10-20T21:34:48}} & \texttt{\tiny{2020-10-20T21:35:48}} \\
    \includegraphics[height=24mm]{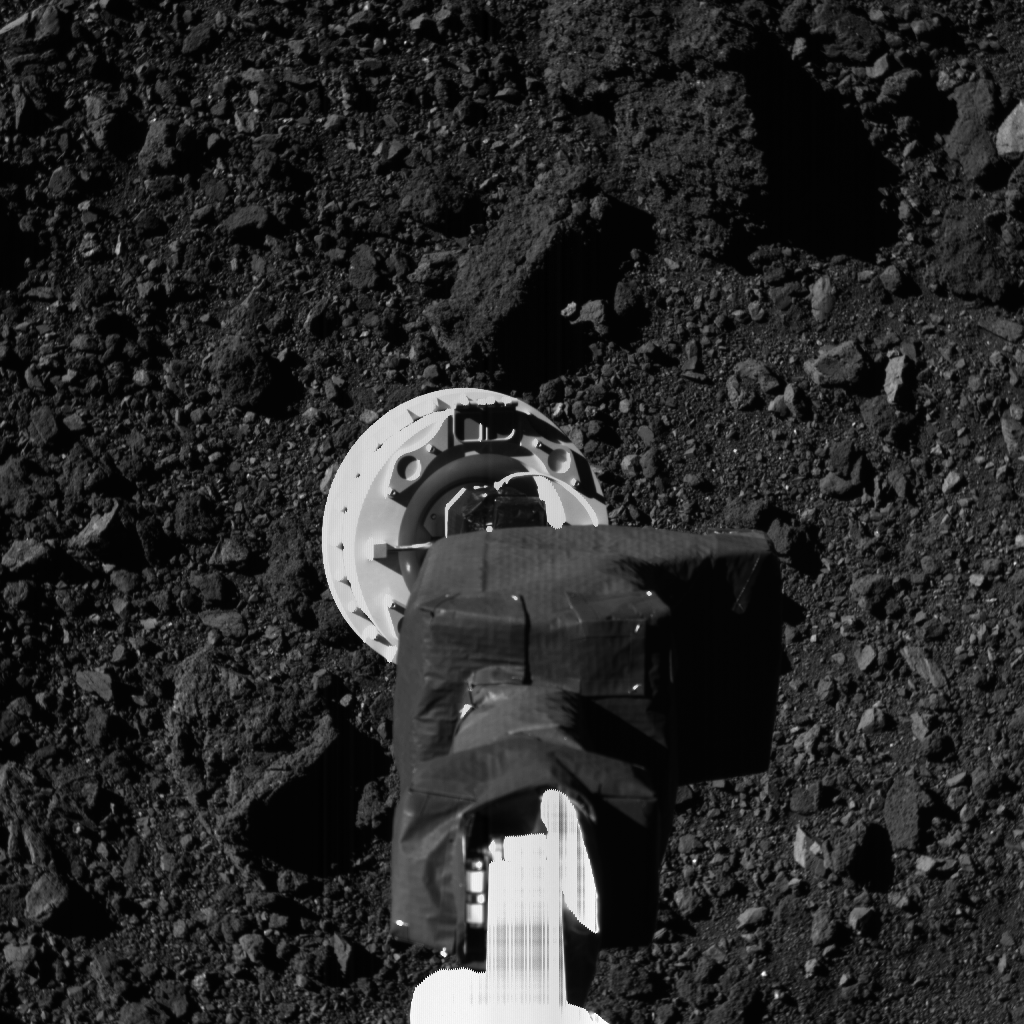} &
    \includegraphics[height=24mm]{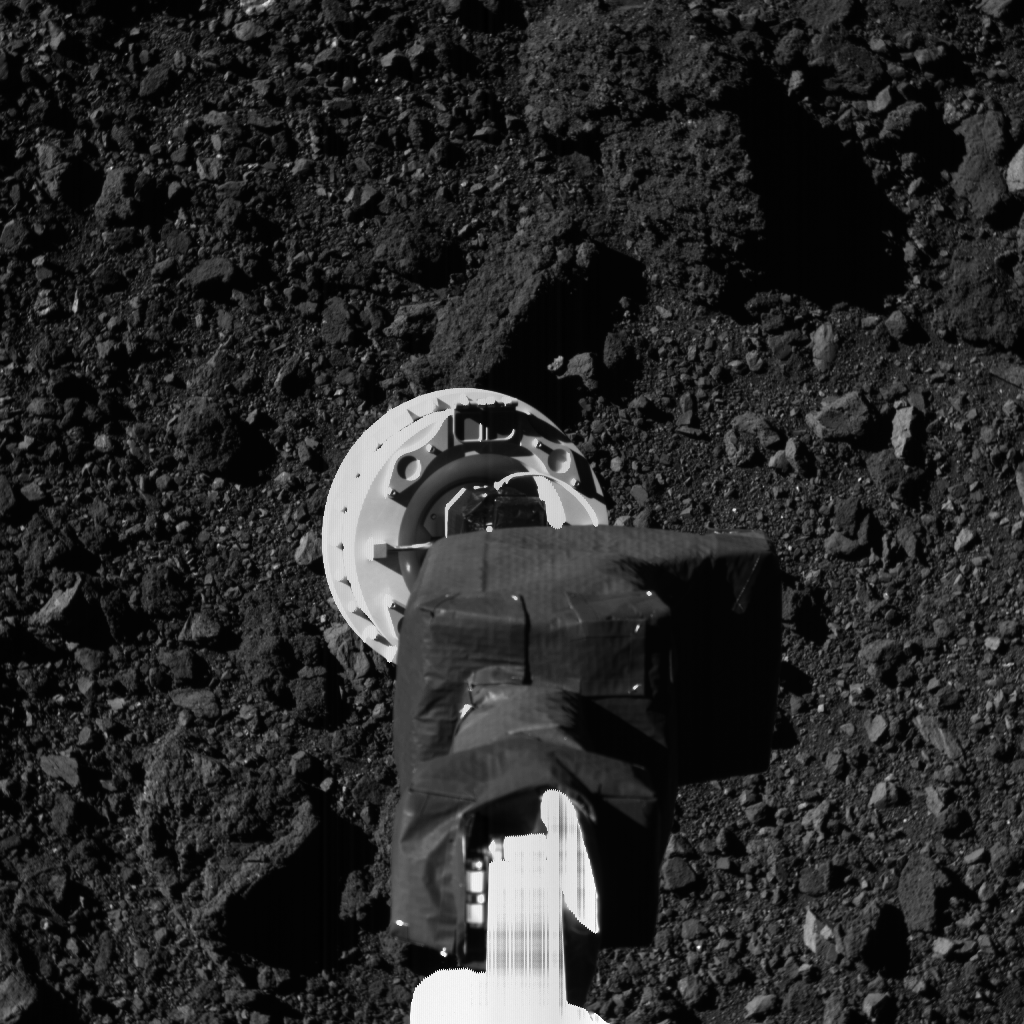} &
    \includegraphics[height=24mm]{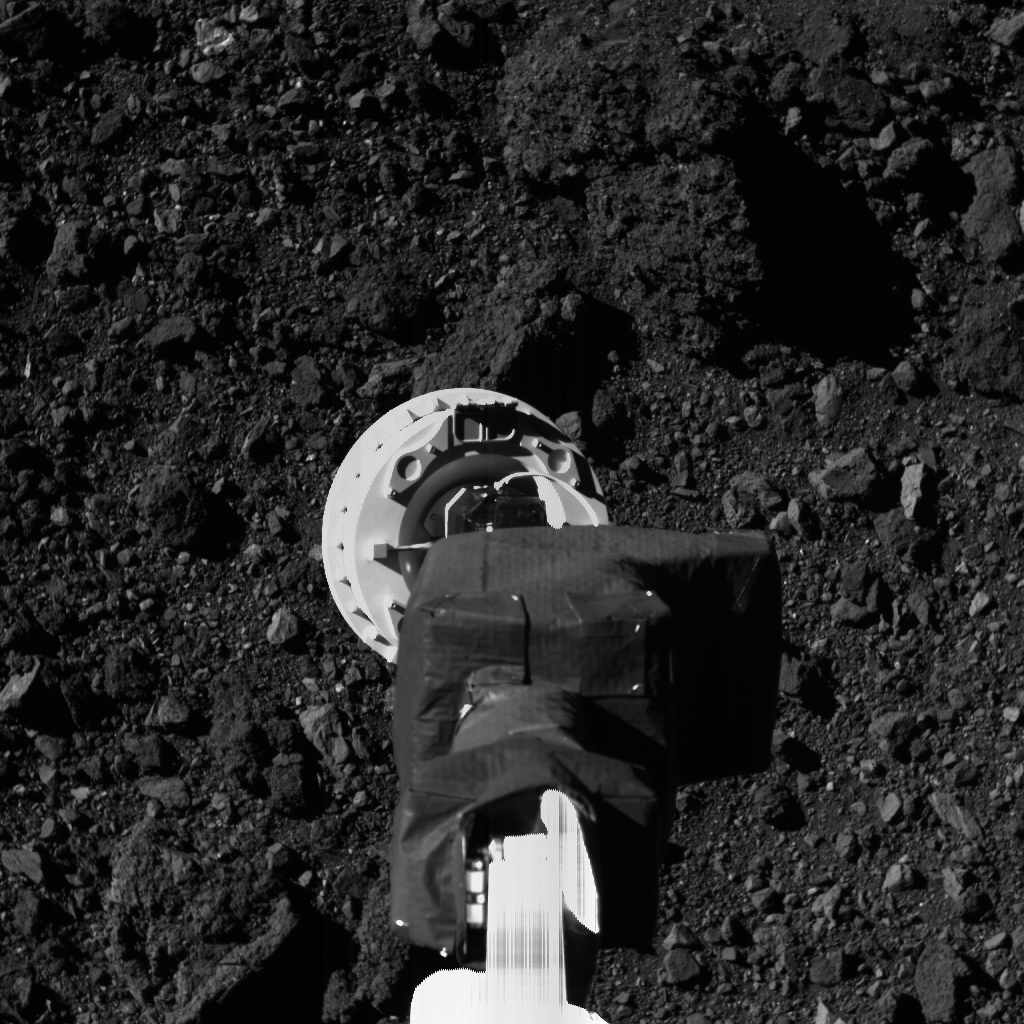} &
    \includegraphics[height=24mm]{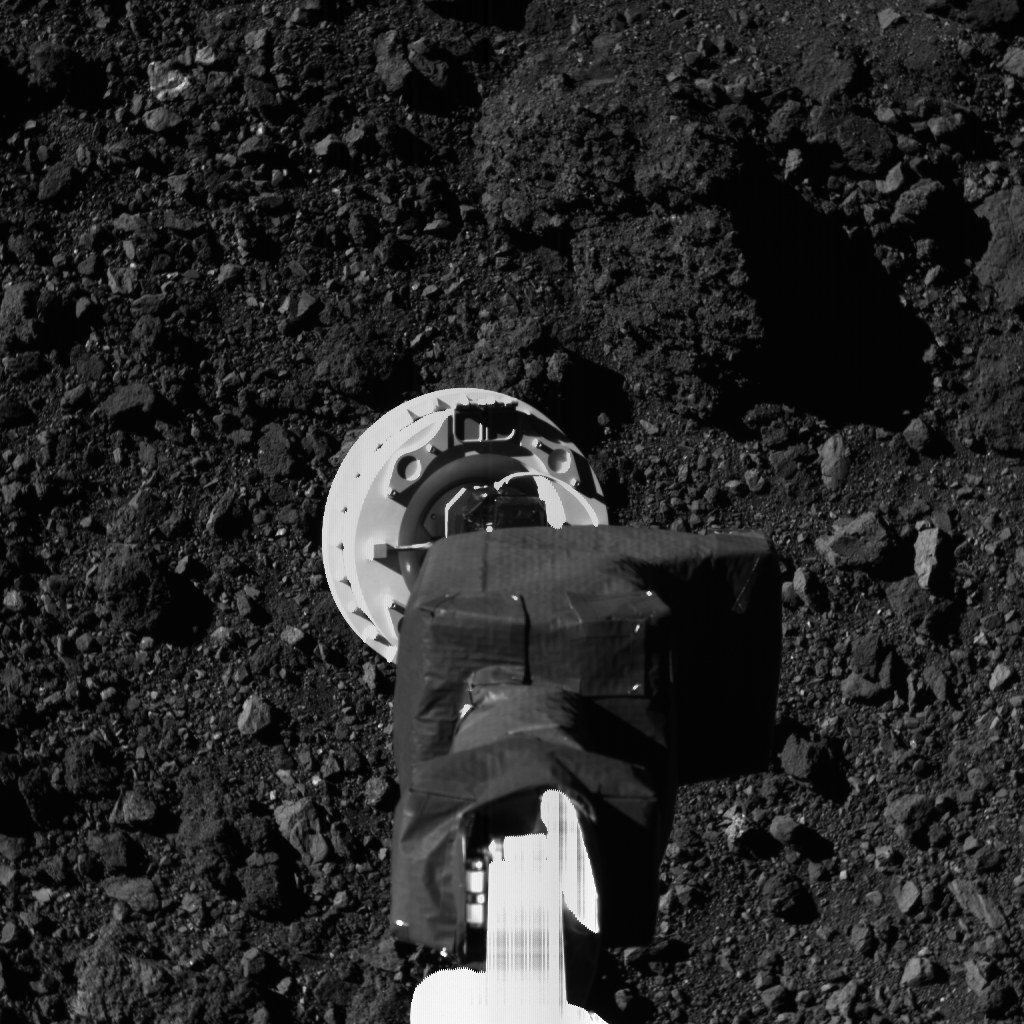} &
    \includegraphics[height=24mm]{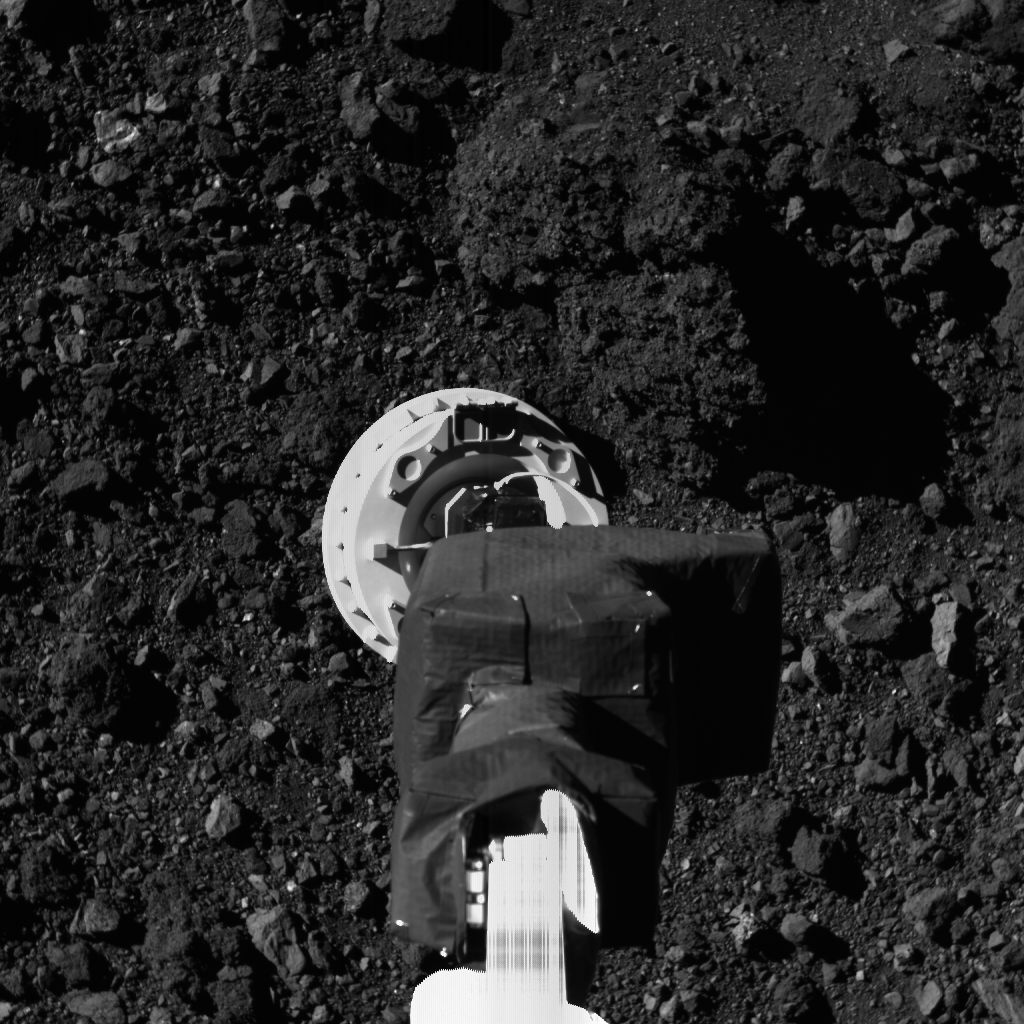} &
    \includegraphics[height=24mm]{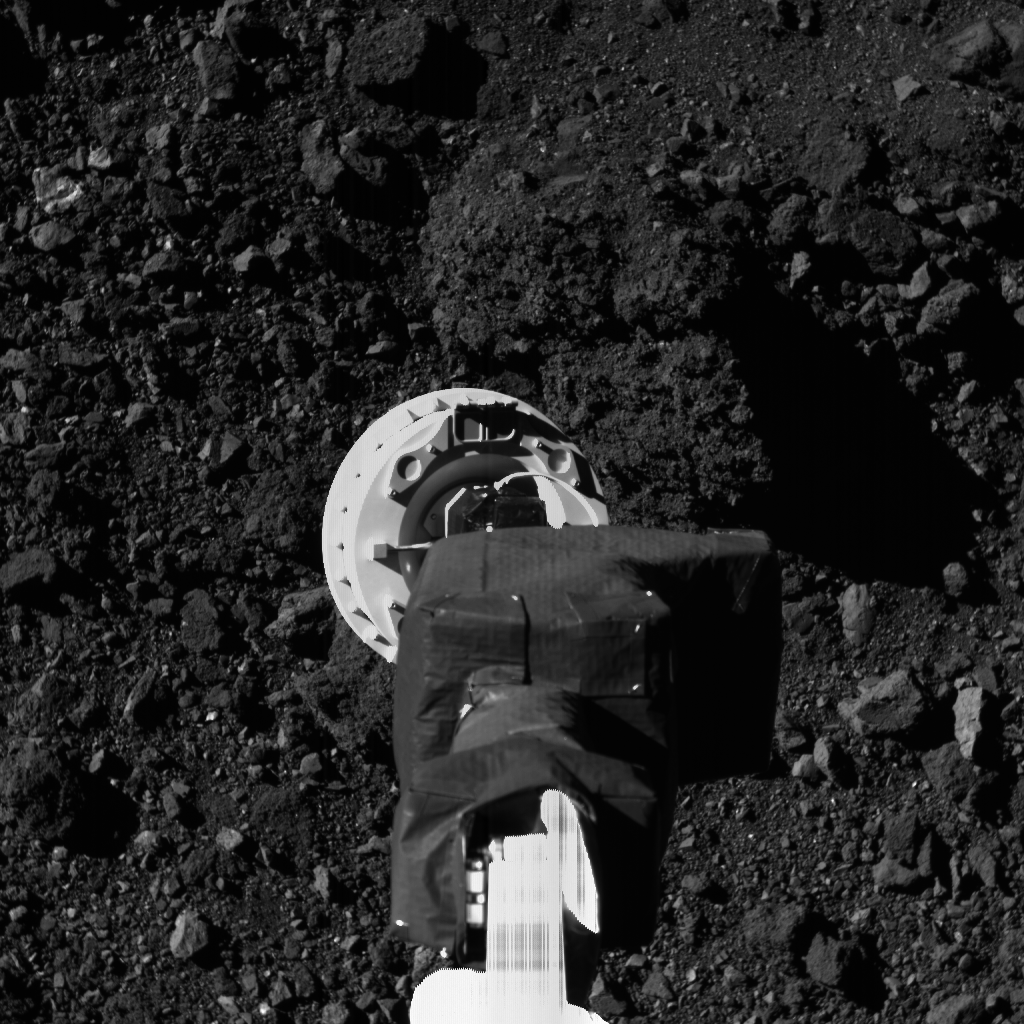} \\
    \texttt{\tiny{2020-10-20T21:36:48}} & \texttt{\tiny{2020-10-20T21:37:48}} & \texttt{\tiny{2020-10-20T21:38:48}} & \texttt{\tiny{2020-10-20T21:39:48}} & \texttt{\tiny{2020-10-20T21:40:48}} & \texttt{\tiny{2020-10-20T21:41:48}} \\
    \includegraphics[height=24mm]{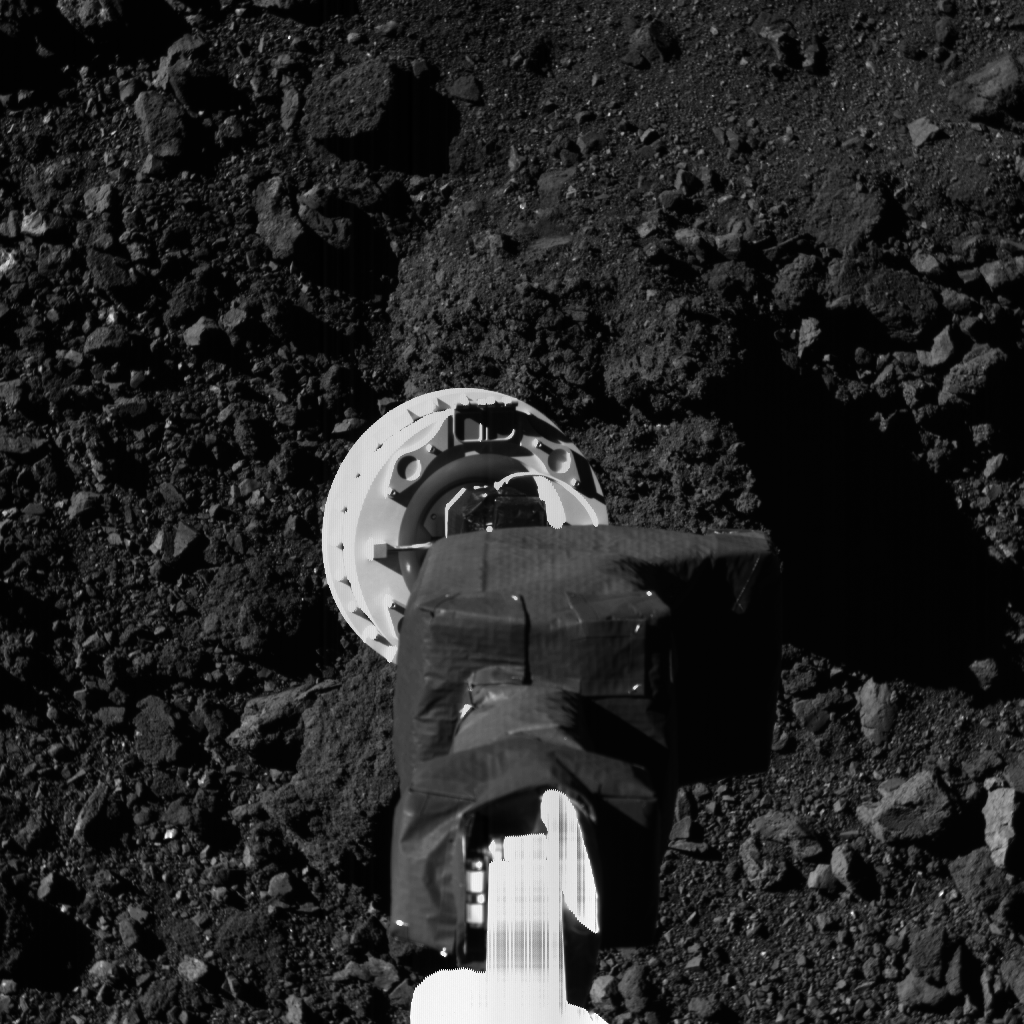} &
    \includegraphics[height=24mm]{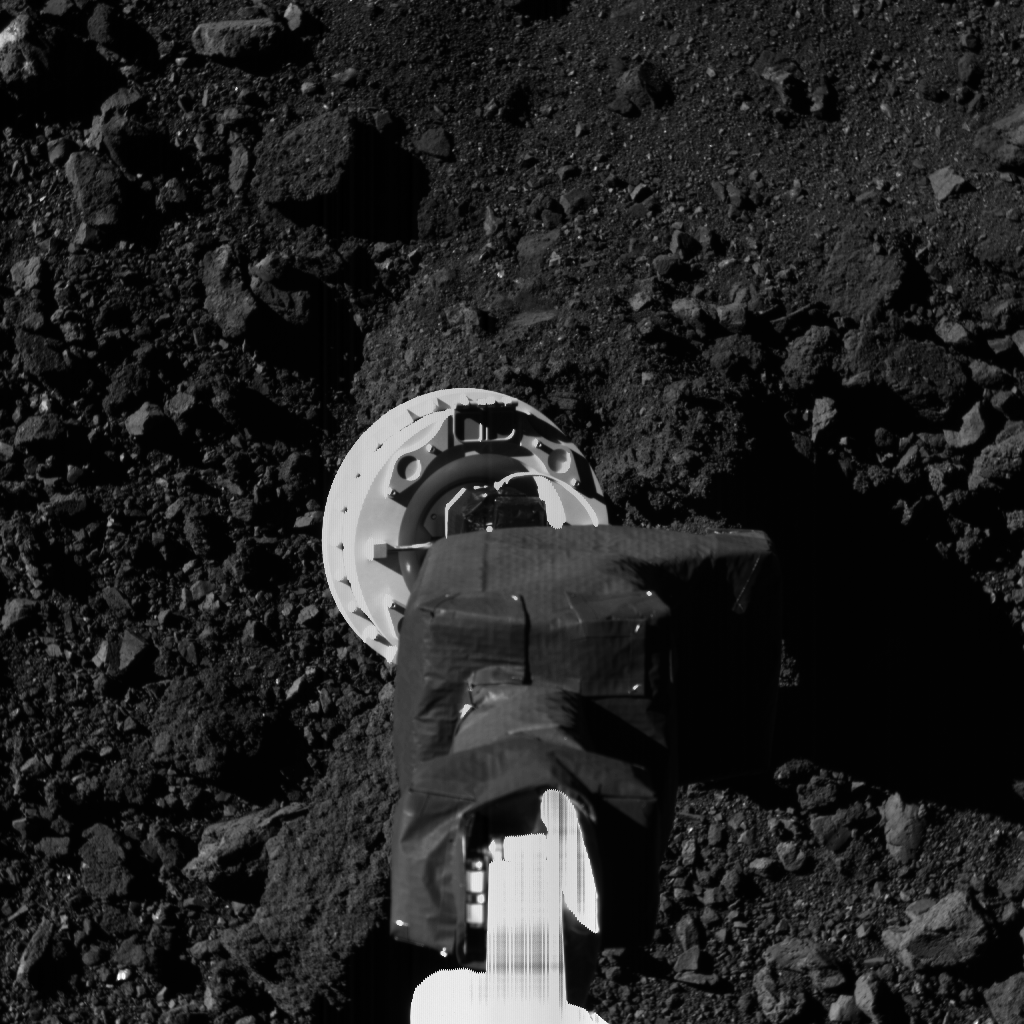} &
    \includegraphics[height=24mm]{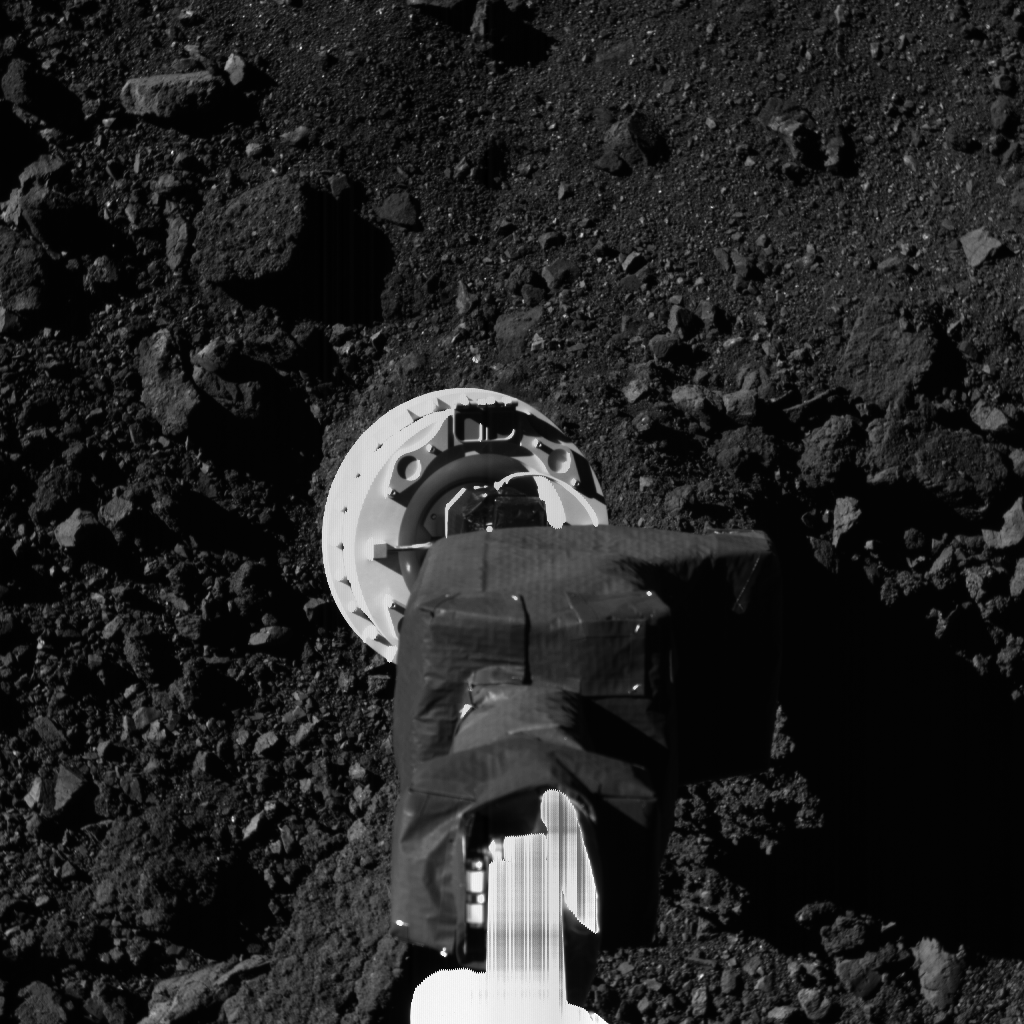} &
    \includegraphics[height=24mm]{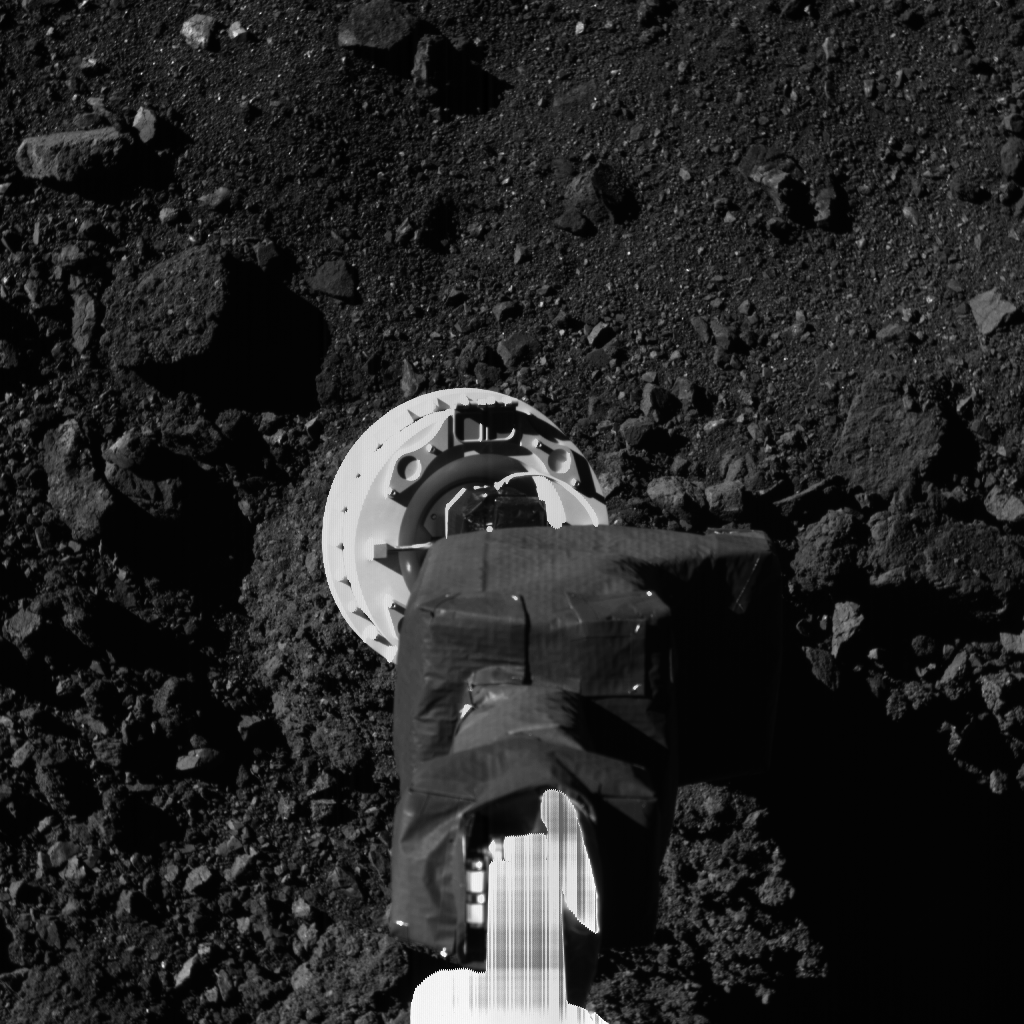} &
    \includegraphics[height=24mm]{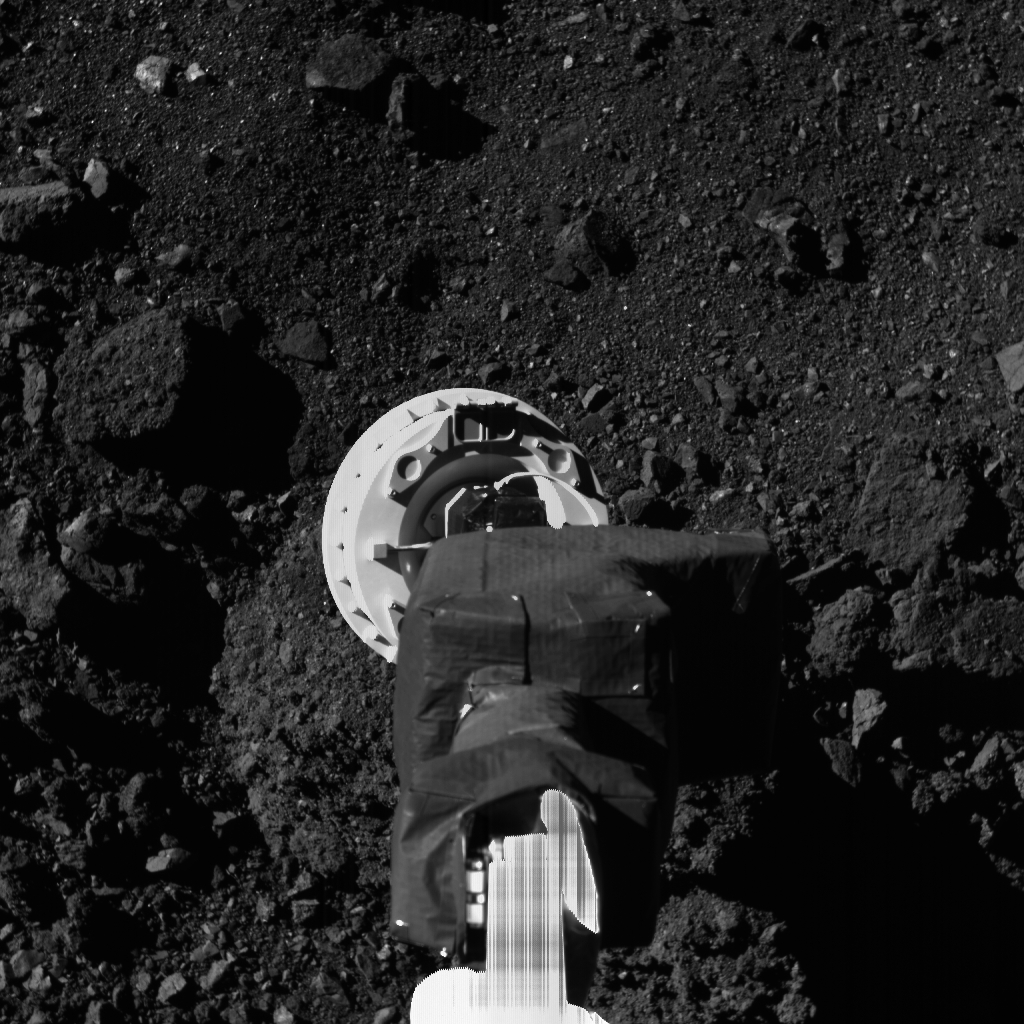} &
    \includegraphics[height=24mm]{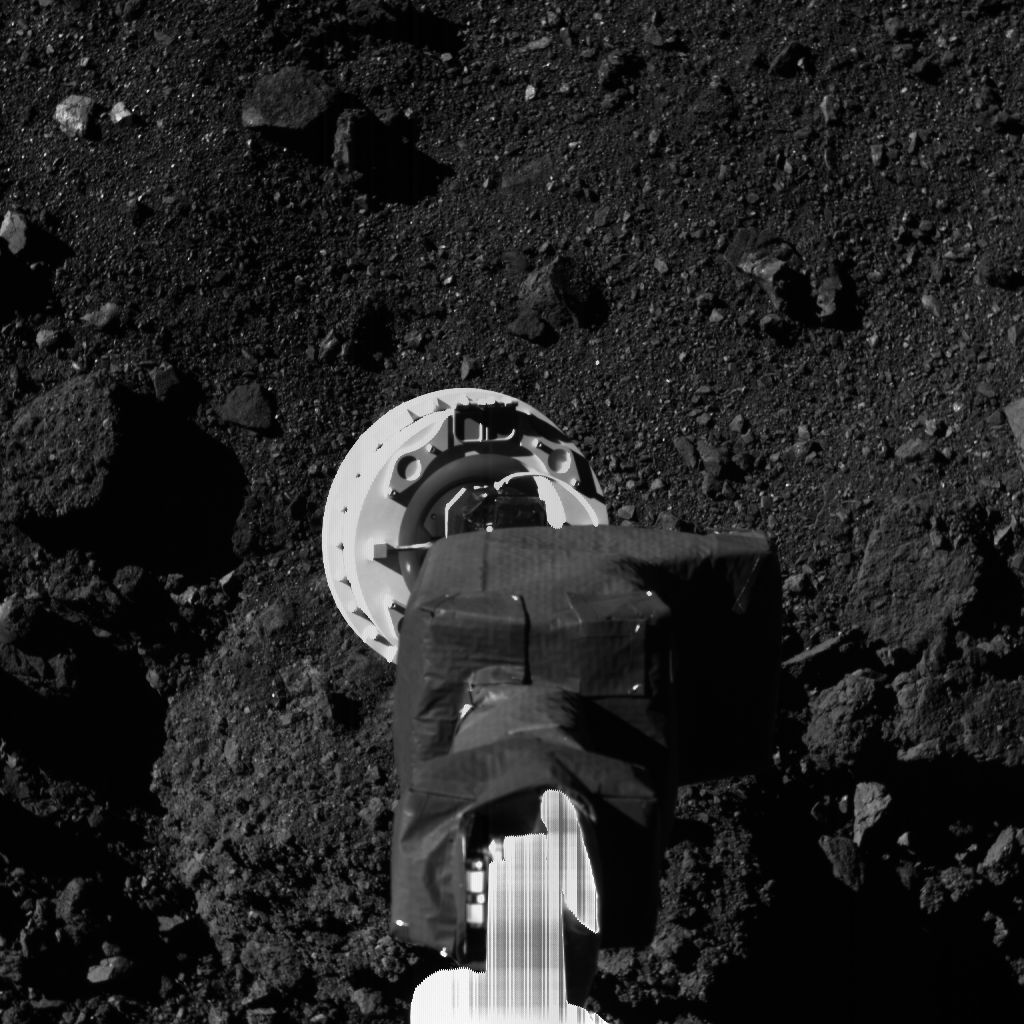} \\
\end{tabular}
\end{center}
\end{minipage}
    \caption{\textbf{Frames from the OSIRIS-Rex TAG sequence captured by the SamCam.}}
    \label{fig:tag-sequence}
\end{figure}

For the second experiment, we trained two models using different combinations of images from the OSIRIS-REx mission: one model, denoted by BICNet-NKO, is trained on Nightingale, Kingfisher, and Osprey, and the other model, denoted by BICNet-KOS, is trained on Kingfisher, Osprey, and Sandpiper. 
These models were trained for the slope and roughness case only. 
We tested the two models on images captured during the TAG sample collection event at the Nightingale sample site. 
We present both models to illustrate the effect of the different training data distributions on the test results. 
A subset of the 42 image sequence is shown in Figure \ref{fig:tag-sequence}.  
Note that the 42 test images are not included in the training set for the Nightingale images.

\begin{figure}[htbp!]
    \begin{minipage}{0.90\linewidth}
\begin{center}
\begin{subfigure}[t]{\linewidth}
\begin{tabular}{c@{\hskip 5pt}c@{\hskip 5pt}c@{\hskip 0pt}c}
    \includegraphics[height=34mm]{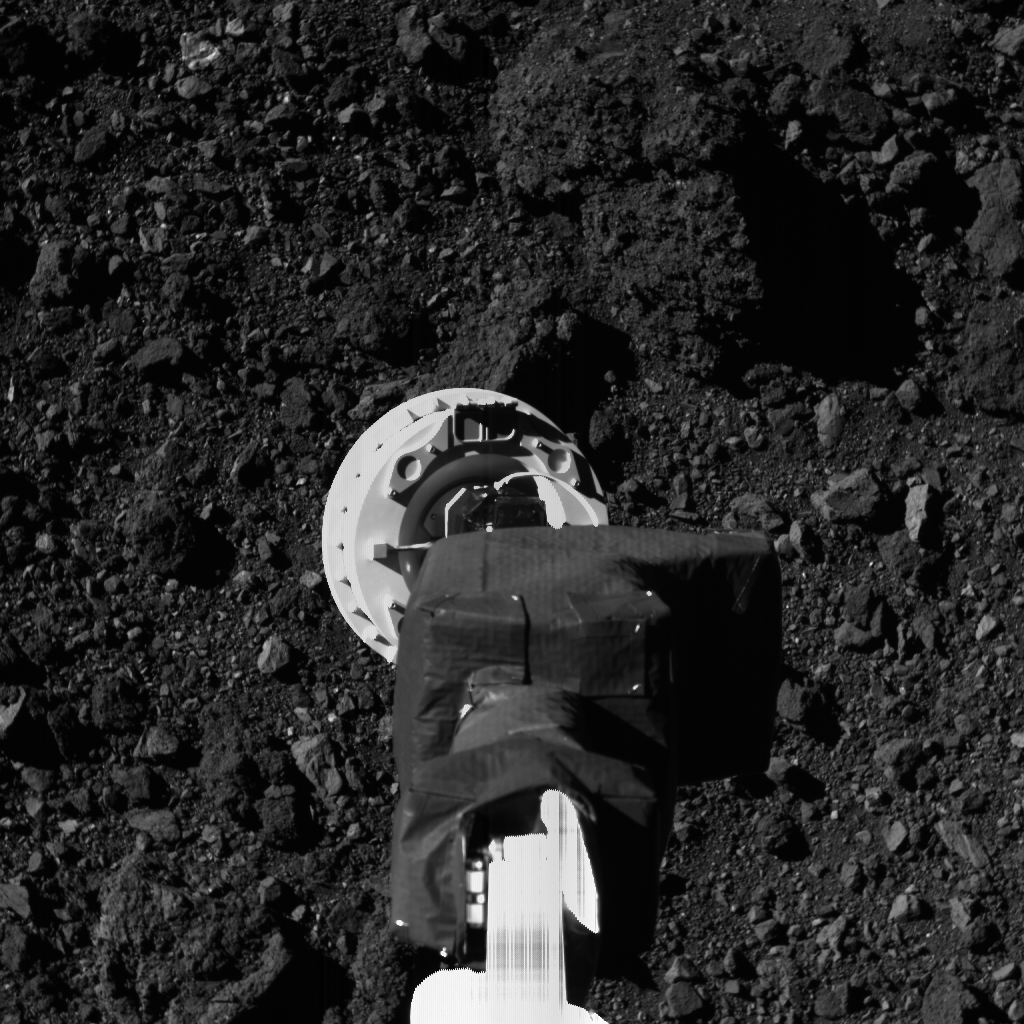} &
    \includegraphics[height=34mm]{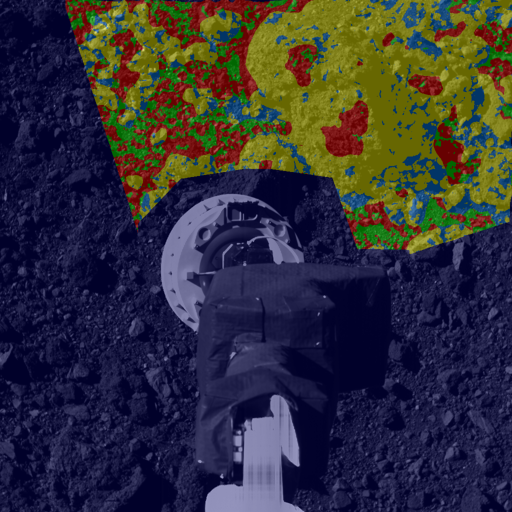} &
    \includegraphics[height=34mm]{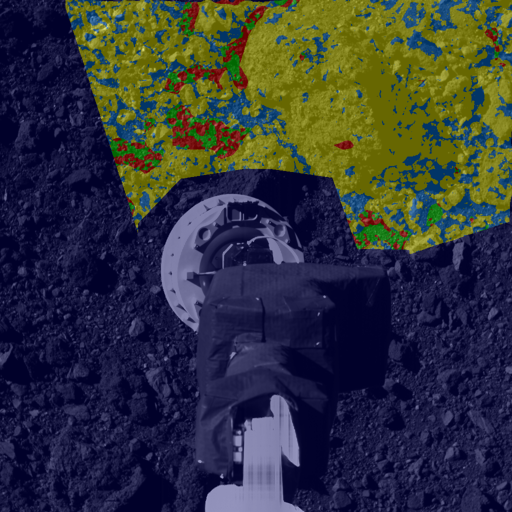} &
    \includegraphics[height=34mm]{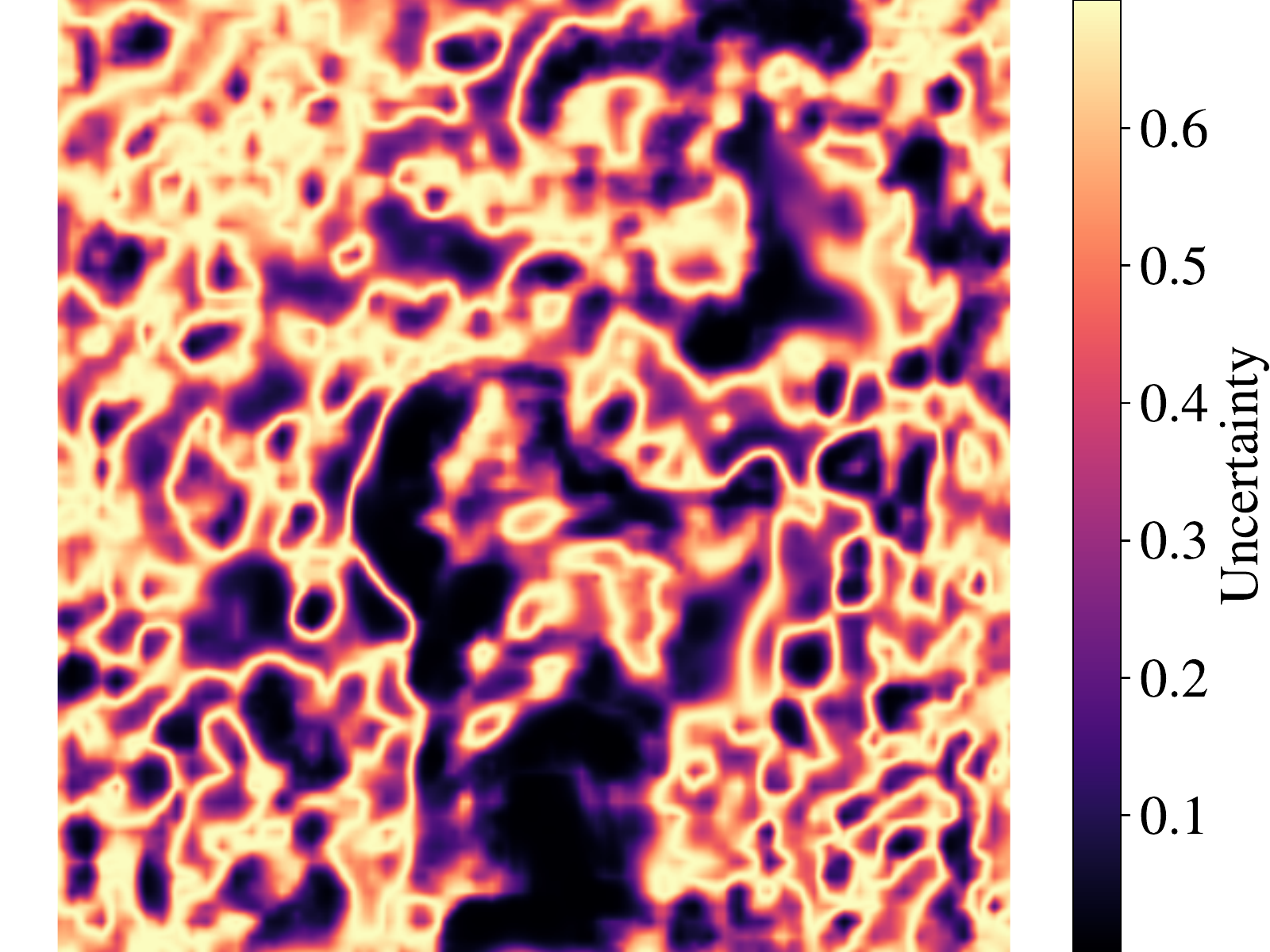} \\
    \includegraphics[height=34mm]{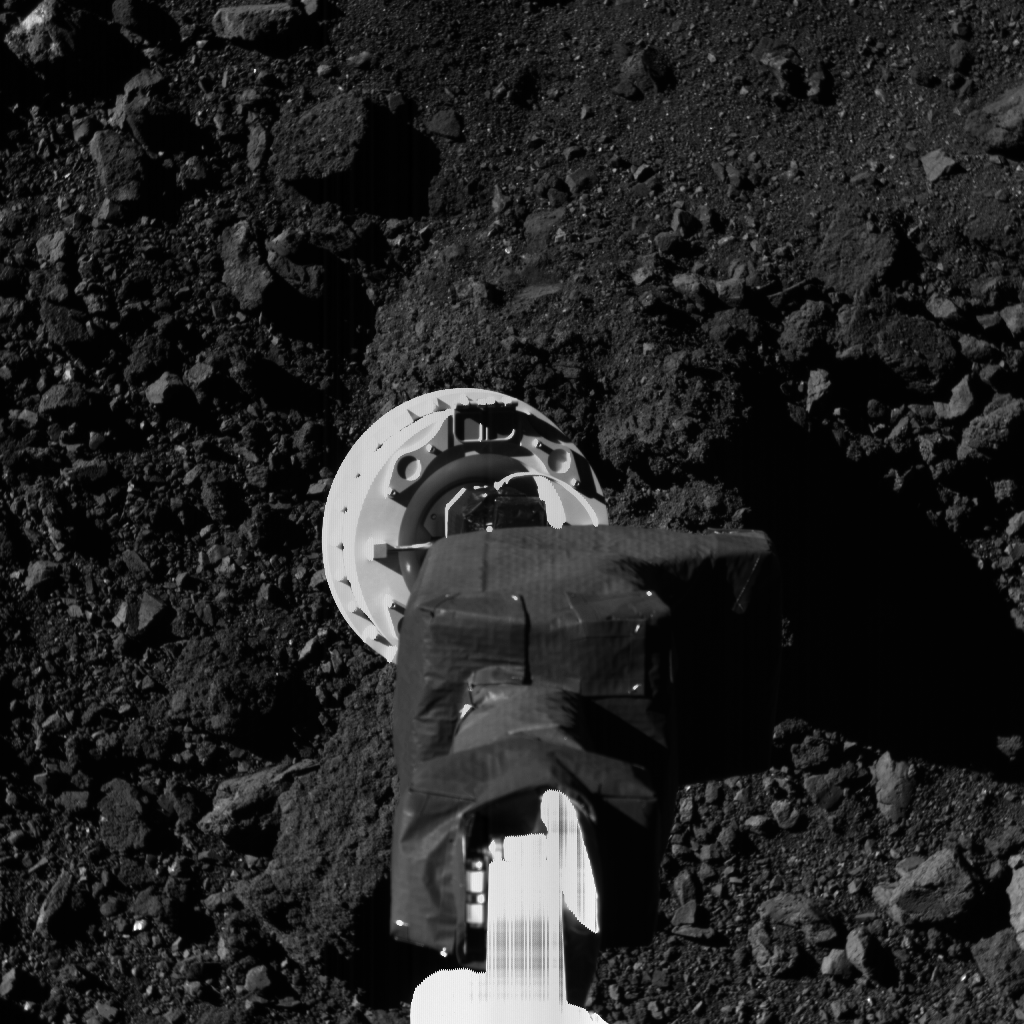} &
    \includegraphics[height=34mm]{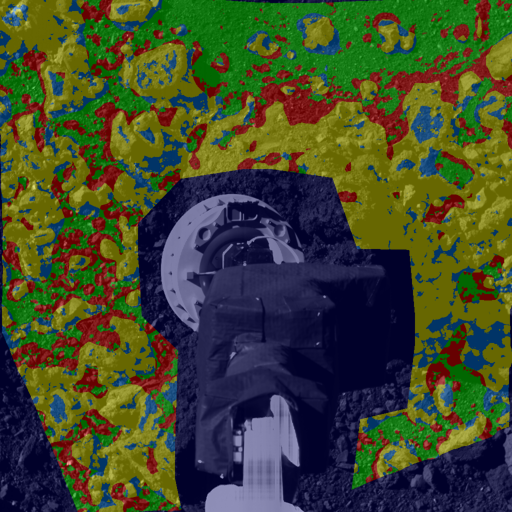} &
    \includegraphics[height=34mm]{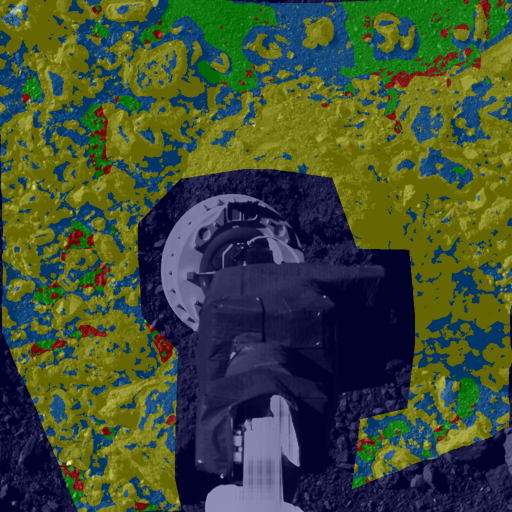} &
    \includegraphics[height=34mm]{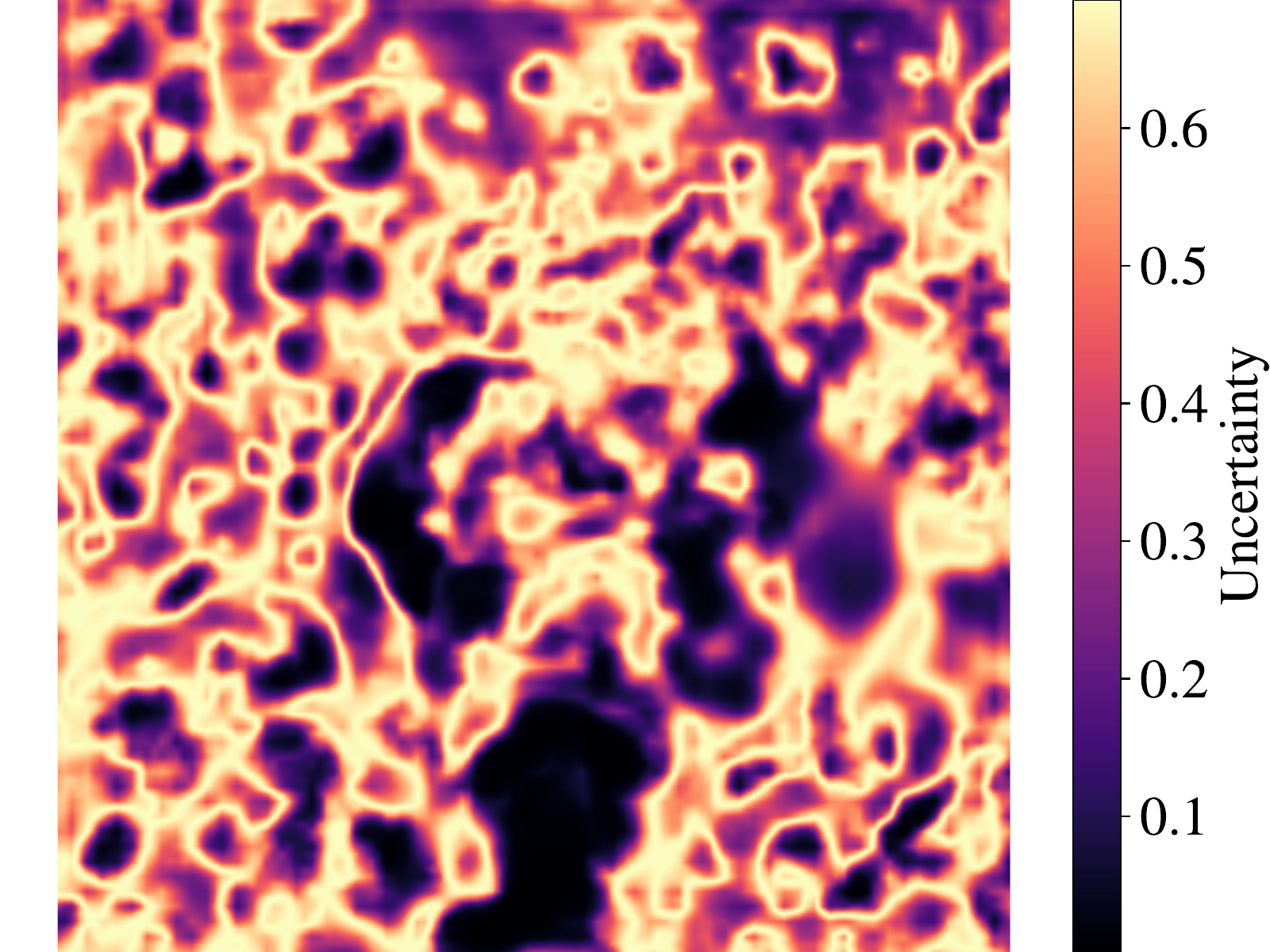} \\
    \includegraphics[height=34mm]{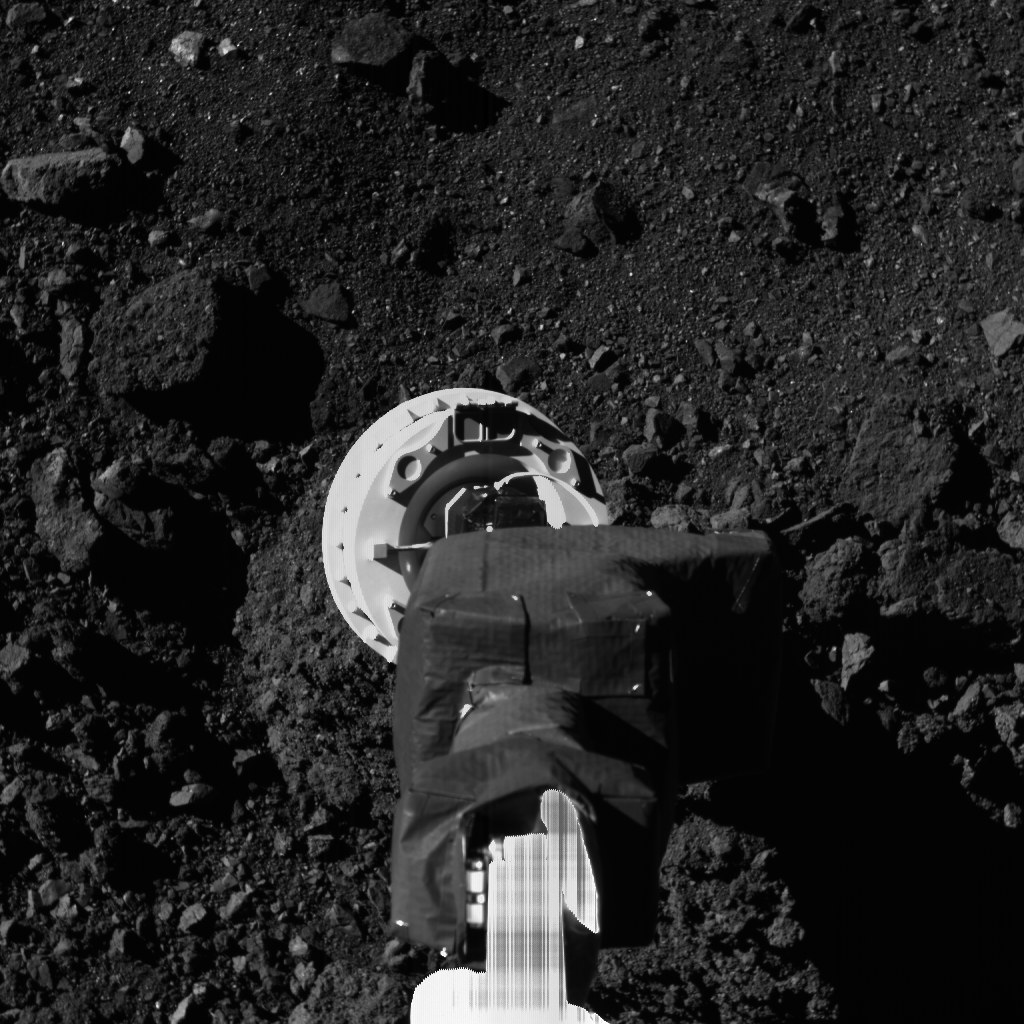} &
    \includegraphics[height=34mm]{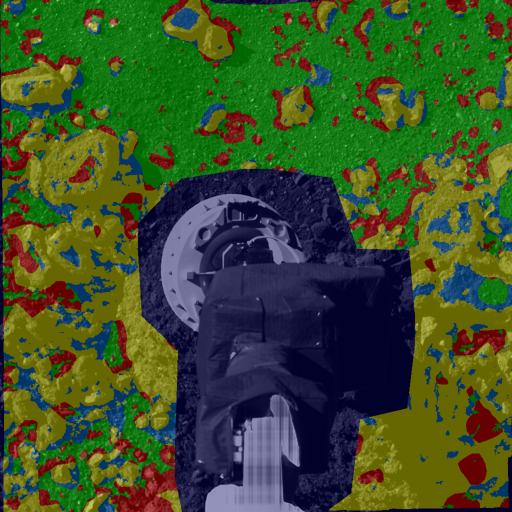} &
    \includegraphics[height=34mm]{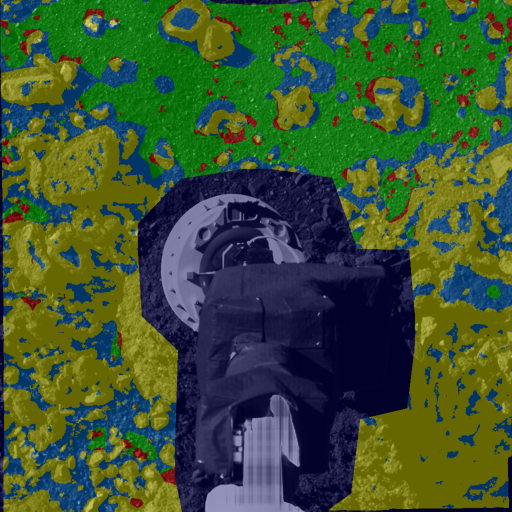} &
    \includegraphics[height=34mm]{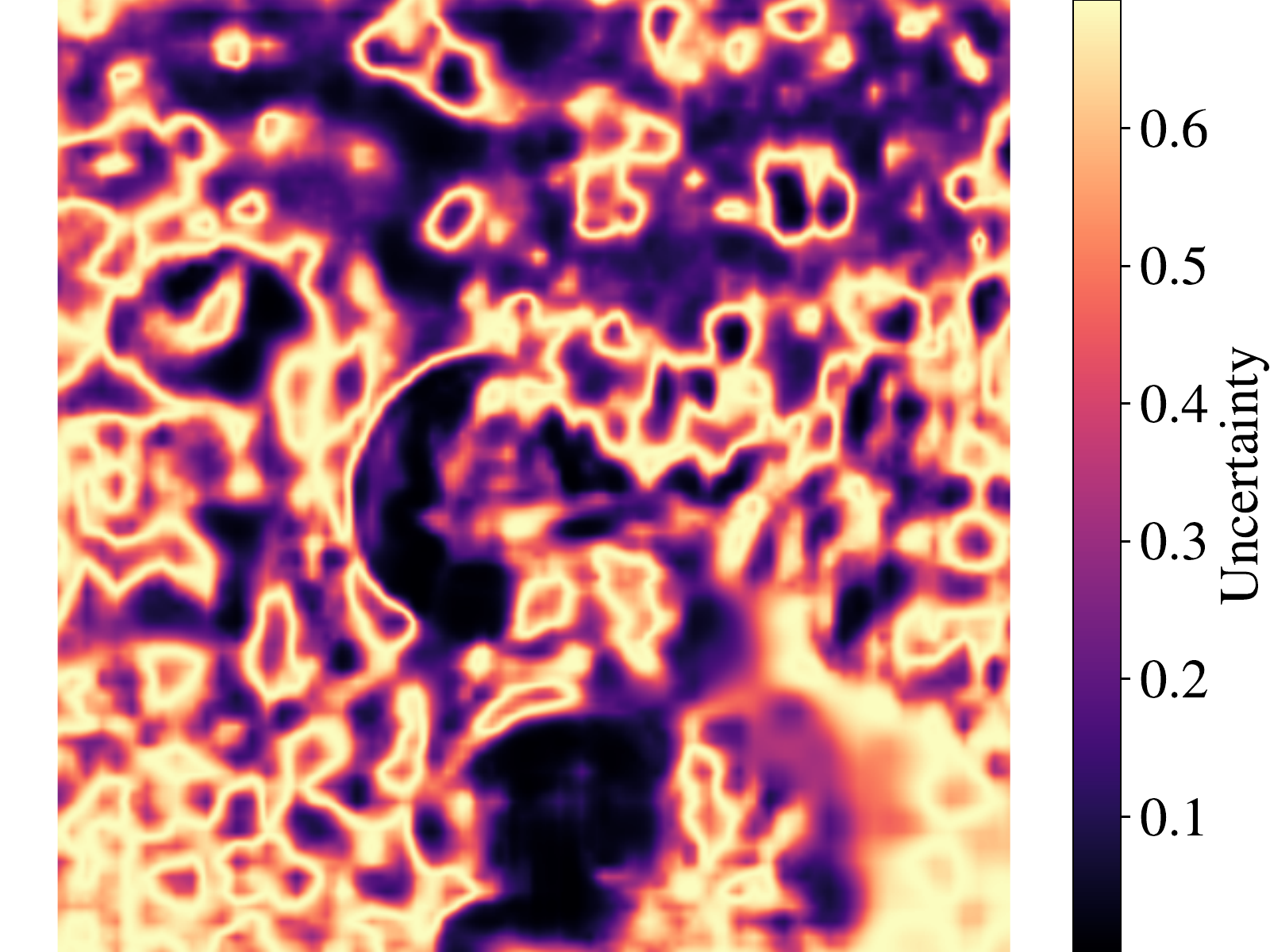} \\
\end{tabular}
\vspace{-5pt}
\caption{BICNet-NKO}
\end{subfigure}
\begin{subfigure}[t]{\linewidth}
\begin{tabular}{c@{\hskip 5pt}c@{\hskip 5pt}c@{\hskip 0pt}c}
    \includegraphics[height=34mm]{Figures/qualitative-tag-kos/00000015.png} &
    \includegraphics[height=34mm]{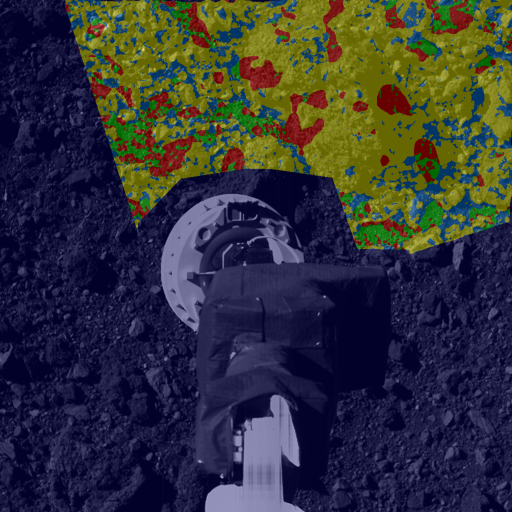} &
    \includegraphics[height=34mm]{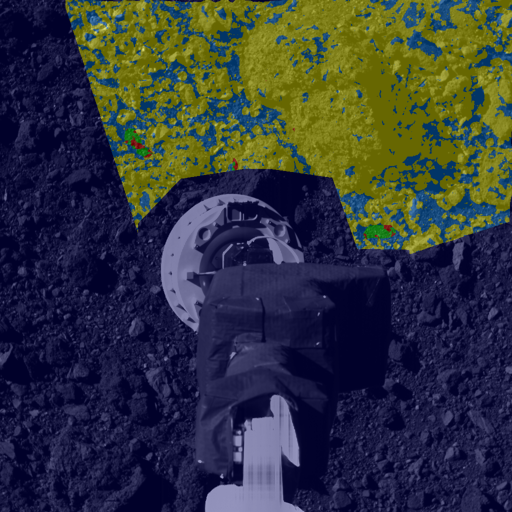} &
    \includegraphics[height=34mm]{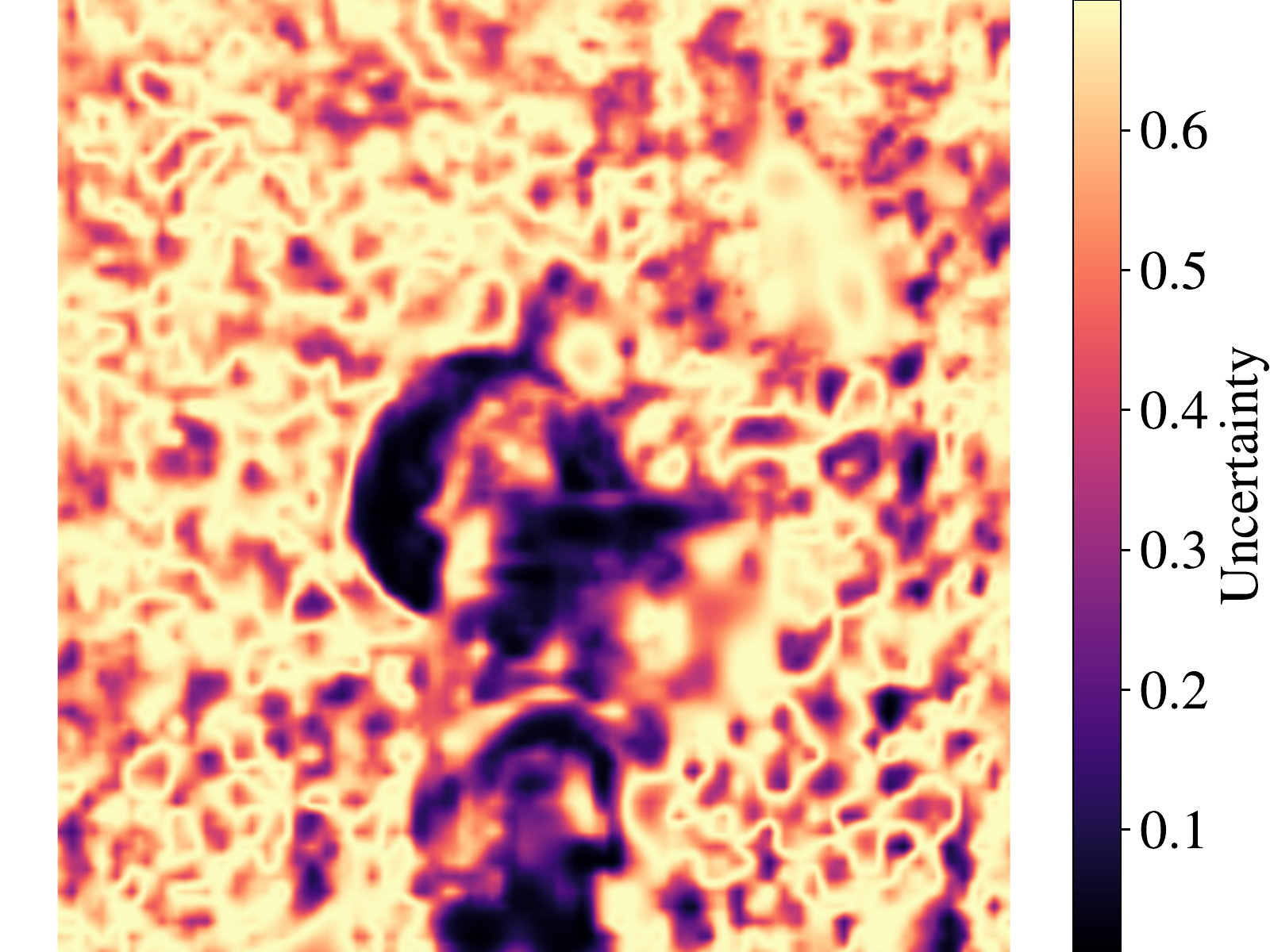} \\
    \includegraphics[height=34mm]{Figures/qualitative-tag-kos/00000028.png} &
    \includegraphics[height=34mm]{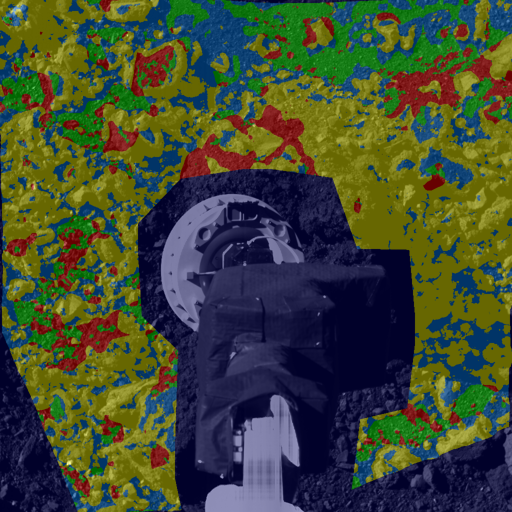} &
    \includegraphics[height=34mm]{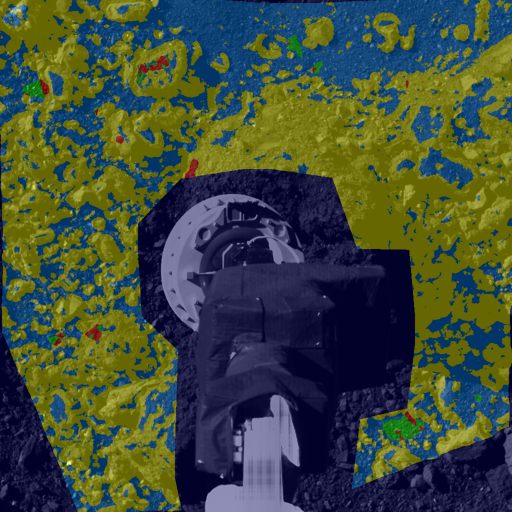} &
    \includegraphics[height=34mm]{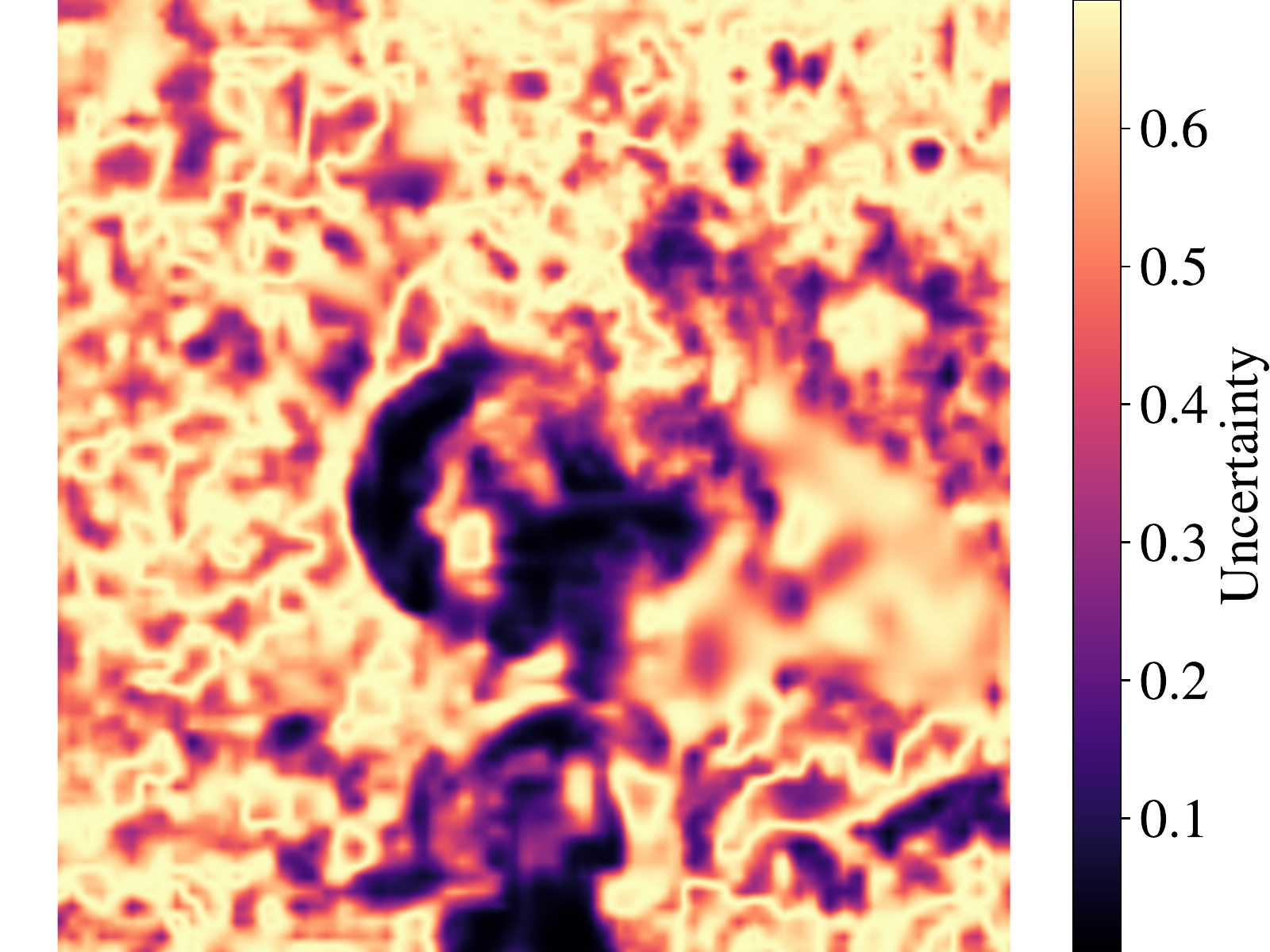} \\
    \includegraphics[height=34mm]{Figures/qualitative-tag-kos/00000036.png} &
    \includegraphics[height=34mm]{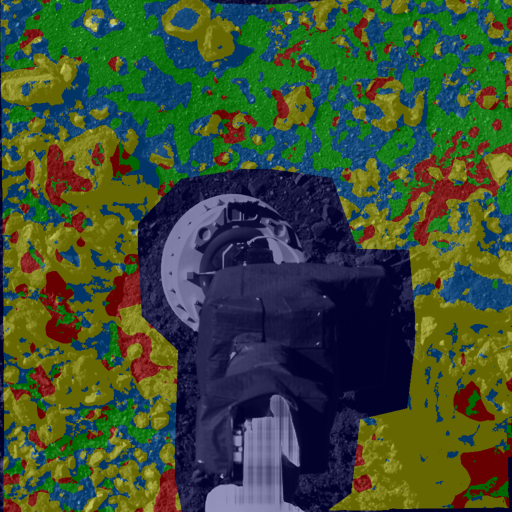} &
    \includegraphics[height=34mm]{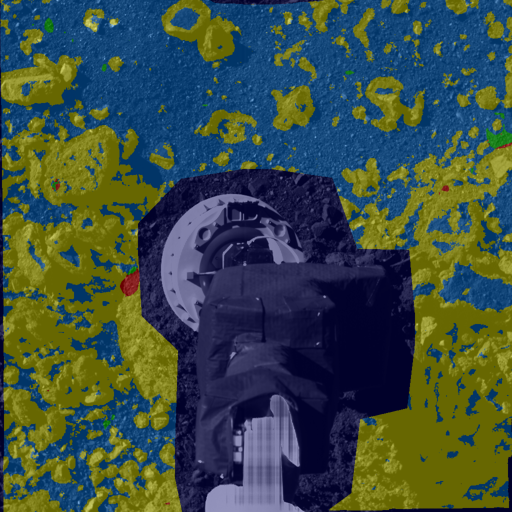} &
    \includegraphics[height=34mm]{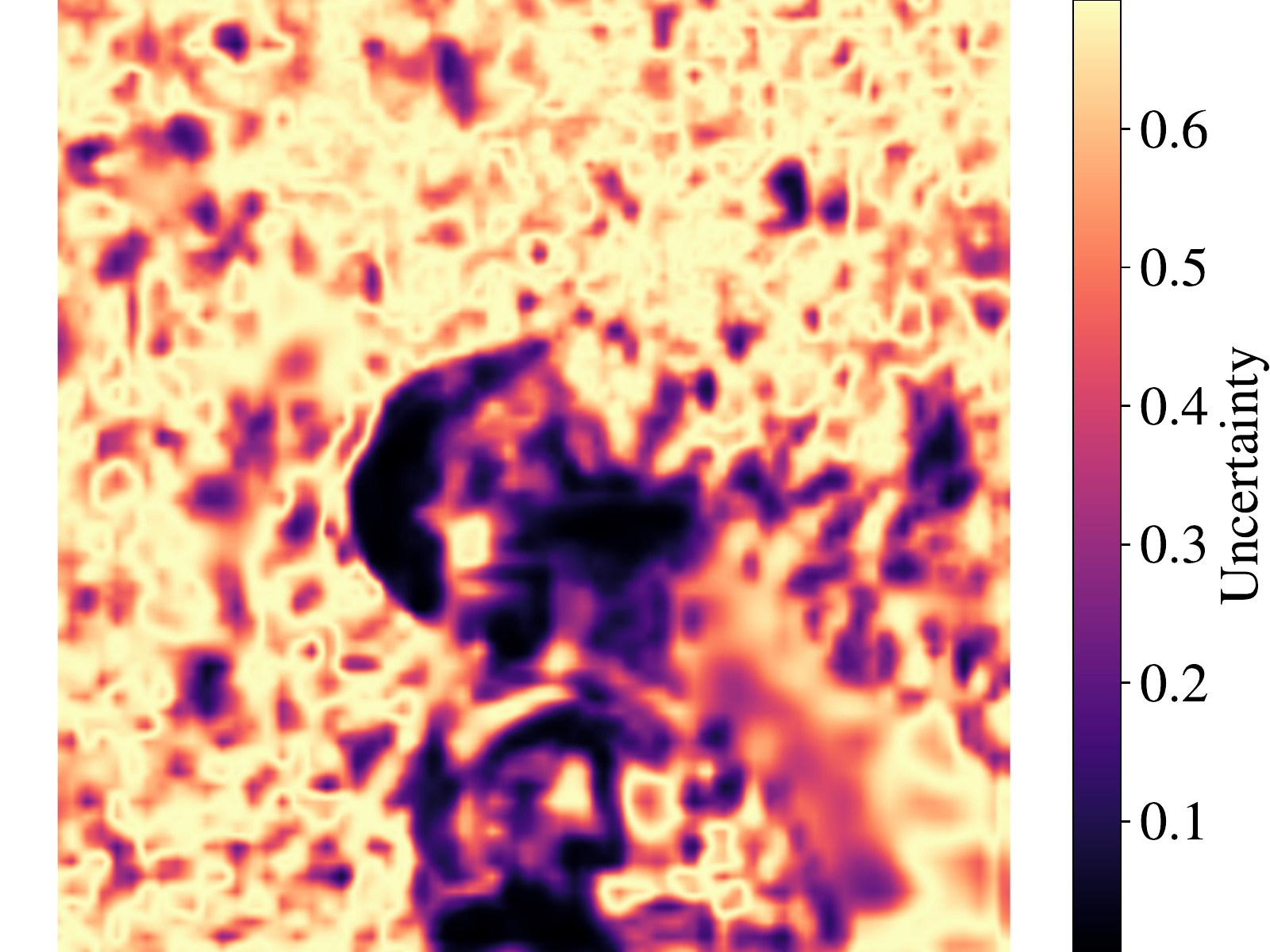} \\
    \footnotesize{Test Image} & \footnotesize{Without Uncertainty} & \footnotesize{With Uncertainty} & \footnotesize{Uncertainty} \\
\end{tabular}
\vspace{-5pt}
\caption{BICNet-KOS}
\end{subfigure}
\end{center}
\end{minipage}
    \vspace{-5pt}
    \caption{\textbf{Qualitative monocular safety mapping results for the TAG experiment.} Green, yellow, blue, and red labels represent true safe, true unsafe, false unsafe, and false safe, respectively. }
    \label{fig:qual-tag-safety}
\end{figure}

Table \ref{tab:safety-metrics-all-tag} and Figure \ref{fig:qual-tag-safety} show quantitative results and qualitative examples, respectively. 
Comparing BICNet-NKO and BICNet-KOS in Table \ref{tab:safety-metrics-all-tag}, we can see sensitivity of BICNet-KOS is significantly lower than that of BICNet-NKO after uncertainty thresholding. 
Indeed, BICNet-KOS assigns a high uncertainty to almost all regions of the input images and are thus overwritten as unsafe after uncertainty thresholding, as shown in Figure \ref{fig:qual-tag-safety}, which is not the case for BICNet-NKO.
These differences are most likely explained by the difference between the training and testing data distributions for BICNet-KOS, as illustrated in Figure \ref{fig:data-distributions}. 
Specifially, the TAG images have less overlap with the images of the prospective landing sites except for Nightingale with respect to viewing angle and visibility ratio. 
Therefore, the exclusion of Nightingale from training data increases the predictive uncertainty for TAG images at test time for BICNet-KOS. 
Note that the Nightingale images (excluding the TAG images) were used for validation during training for the BICNet-KOS model in order to rule out overfitting as a cause for the decreased prediction performance. 

\begin{figure}
    \centering
    \begin{subfigure}[c]{\linewidth}
        \includegraphics[width=\linewidth]{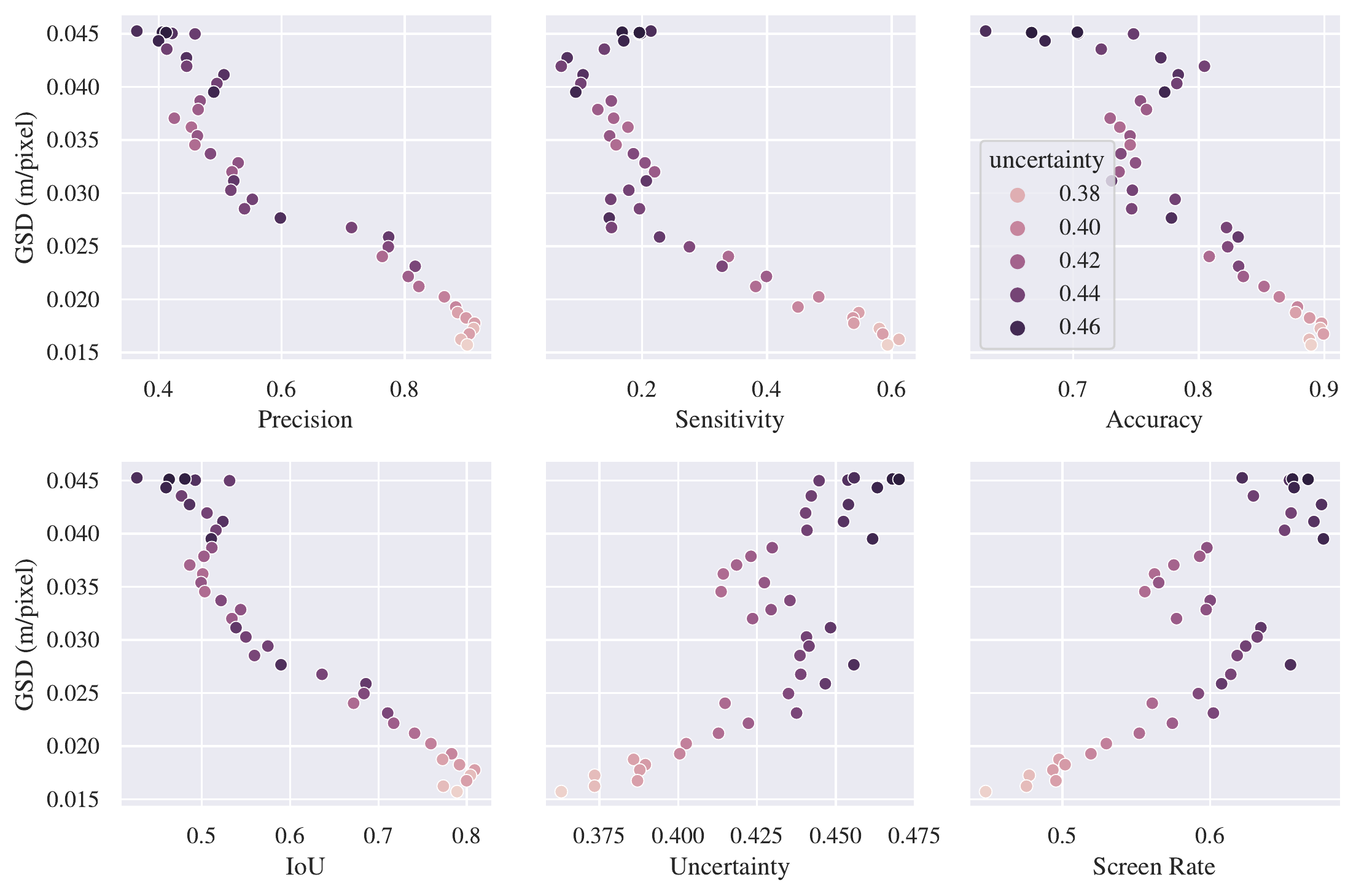}
        \caption{BICNet-NKO}
    \end{subfigure}\\
    \begin{subfigure}[c]{\linewidth}
        \includegraphics[width=\linewidth]{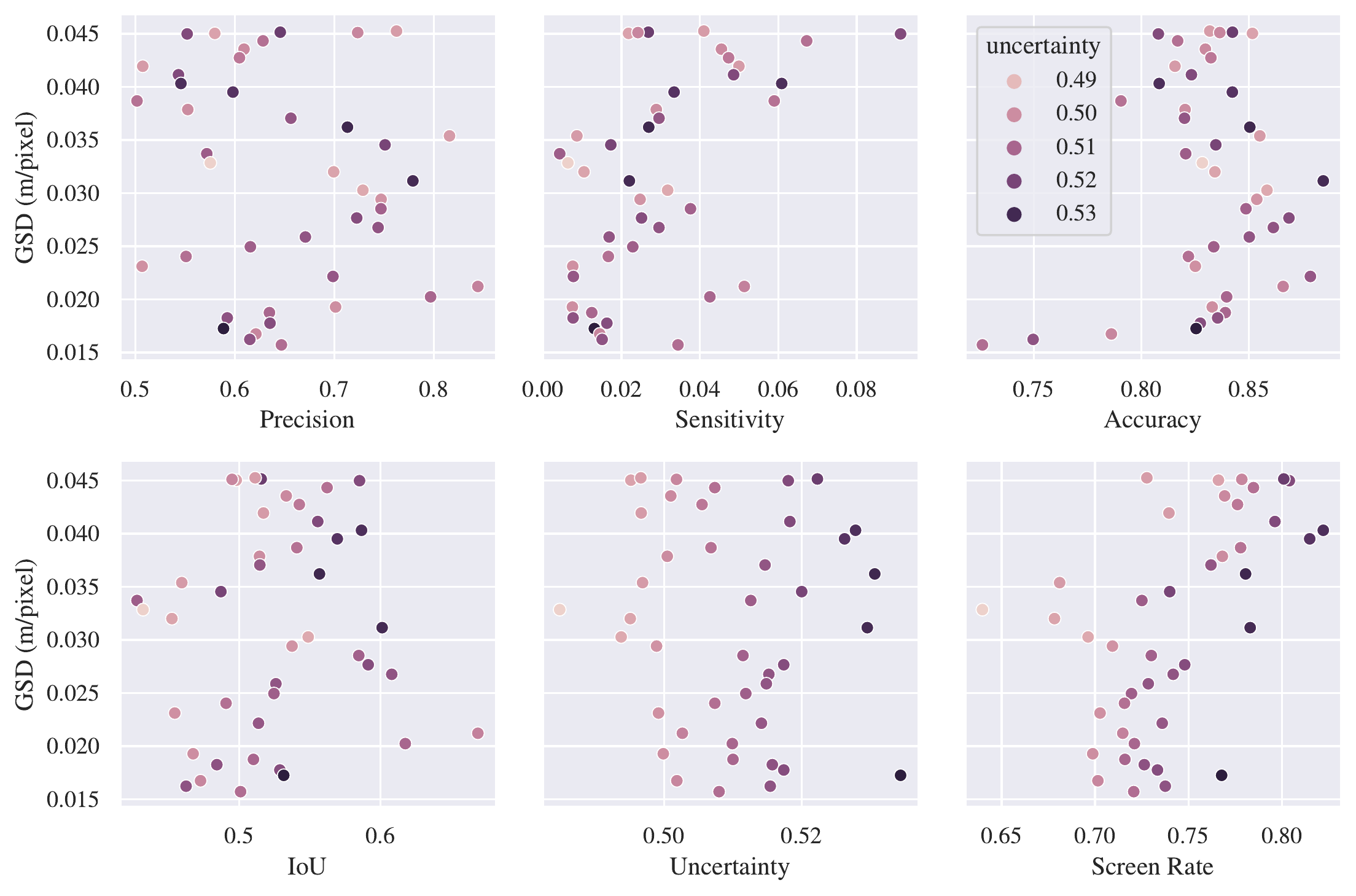}
        \caption{BICNet-KOS}
    \end{subfigure}
    \caption{\textbf{Per-image metrics for the TAG experiment with respect to the GSD for \textit{slope \& roughness} safety.}}
    \label{fig:eval-tag-slope-rghns-gsd}
\end{figure}

This effect of training data distributions on the uncertainty level, and the accompanying predictive performance, is consistent with the per-image metrics with respect to GSD, as shown in Figure \ref{fig:eval-tag-slope-rghns-gsd}.
Indeed, for BICNet-NKO, as the GSD of testing images gets closer to the peak of the training data, the predictive performance increases and uncertainty decreases. 
Lower precision for the higher GSD images is also partly due to the very low incidence of safe regions for these instances (see Figure \ref{fig:qual-tag-safety}). 
Conversely, for BICNet-KOS, all test images have a high screening rate and a low sensitivity due to the high uncertainty, purportedly due to the out-of-distribution training data with respect to the viewing angle and visibility ratio and the relatively small size of the training set (386 images). 
We do not provide the per-image metrics with respect to the viewing angle and visibility ratio, as these values remain relatively constant over the entire TAG sequence at $\sim$$7^\circ$ and $\sim$$71\%$, respectively, as shown in Figure \ref{fig:data-distributions}.
These results illustrate the effect of training data distributions and the ability of the uncertainty measure to identify out-of-distribution data for uncertainty-aware segmentation networks. 
We postulate that training our model with a more comprehensive set of images will decrease prediction uncertainty and increase the performance. 

\begin{table*}[tp!]
\footnotesize
\centering
\scshape
\ra{1.5}
\caption{\textbf{Overall performance for the TAG experiment for slope \& roughness safety.} BICNet-NKO is our Bayesian ICNet model trained on Nightingale, Kingfisher, and Osprey, and BICNet-KOS is trained on Kingfisher, Osprey, and Sandpiper. All reported values are percentages.}
\begin{adjustbox}{width=\linewidth}
\begin{tabular}{llrrrr}
\toprule
Method & & Precision & Sensitivity & Accuracy & mIoU \\
\midrule 
BICNet-NKO & Without uncertainty   & 49.02 (50.21) & \textbf{61.16} (\textbf{66.53}) & 67.44 (67.09) & 48.65 (48.66) \\
           & With uncertainty      & 61.38 (61.66) & 26.60 (30.80) & 78.62 (77.09) & \textbf{60.04} (\textbf{59.23}) \\
\midrule 
\rowcolor[gray]{0.9}
BICNet-KOS & Without uncertainty   & 50.08 (50.42) & 42.18 (50.37) & 67.24 (65.32) & 45.48 (45.11) \\
\rowcolor[gray]{0.9}
           & With uncertainty      & \textbf{65.05} (\textbf{64.89}) & 2.87 (4.05) & \textbf{83.11} (\textbf{82.65}) & 52.58 (55.87) \\
\bottomrule
\end{tabular}
\end{adjustbox}
\label{tab:safety-metrics-all-tag}
\end{table*}

\section{Conclusion}

In this paper we presented a novel landing hazard detection approach for small body missions that predicts safety maps directly from monocular imagery. 
We implemented an efficient, uncertainty-aware segmentation network that demonstrated hazard detection performance at over $80\%$ accuracy and over $85\%$ precision on real images of unseen landing sites captured during the OSIRIS-REx mission to Asteroid 101955 Bennu. 
We believe that monocular safety mapping is a promising technology for reducing reliance on human-in-the-loop procedures used in current safety mapping methodologies. 
Future work will involve developing a more comprehensive distribution of training data and identifying and rectifying causes of uncertainty to increase the reliability of the proposed approach. 
Our code, data, and trained models will be made available to the public at \url{https://github.com/travisdriver/deep_monocular_hd}.


\section{Acknowledgments}

This work supported by a NASA Space Technology Graduate Research Opportunity and the NASA Early Career Faculty Program (grant no. 80NSSC20K0064).
The authors would like to thank Kenneth Getzandanner and Michael Shoemaker from NASA Goddard Space Flight Center for several helpful discussions and comments.


\bibliographystyle{AAS_publication}   
\bibliography{references}   

\end{document}